\documentclass[10pt,journal,compsoc]{IEEEtran}
\usepackage{amsmath,amsfonts,bm}

\def\eqref#1{equation~\ref{#1}}

\def\1{\bm{1}}

\def\vb{{\bm{b}}}

\def\vh{{\bm{h}}}

\def\vs{{\bm{s}}}

\def\vx{{\bm{x}}}

\def\mD{{\bm{D}}}

\def\mX{{\bm{X}}}

\DeclareMathAlphabet{\mathsfit}{\encodingdefault}{\sfdefault}{m}{sl}
\SetMathAlphabet{\mathsfit}{bold}{\encodingdefault}{\sfdefault}{bx}{n}

\usepackage{xcolor}
\usepackage{amsfonts, amsmath, amssymb, amsthm}
\usepackage{graphicx}
\usepackage{subcaption}
\usepackage{float}
\usepackage{tabularx}

\usepackage{booktabs}
\usepackage{setspace}
\usepackage{algpseudocode}
\usepackage{multirow}

\newtheorem*{remark}{Remark}
\usepackage{colortbl}
\usepackage[ruled,vlined,linesnumbered]{algorithm2e}
\usepackage{subcaption}
\usepackage{threeparttable}
\usepackage[colorlinks=true,linkcolor=red,citecolor=blue,urlcolor=blue]{hyperref}
\usepackage{orcidlink}

\usepackage[acronym, nohypertypes={acronym}]{glossaries}

\ifCLASSOPTIONcompsoc
  \usepackage[nocompress]{cite}
\else
  \usepackage{cite}
\fi
\ifCLASSINFOpdf
\else
\fi
\begin{document}
%



\title{Collaborative Imputation of Urban Time Series through Cross-city Meta-learning}

%
%
%
%

\author{Tong Nie\textsuperscript{~\orcidlink{0000-0001-8403-6622}},
Wei Ma\textsuperscript{~\orcidlink{0000-0001-8945-5877}}$^\dagger$~\IEEEmembership{Member,~IEEE}, Jian Sun\textsuperscript{~\orcidlink{0000-0001-5031-4938}}$^\dagger$, Yu Yang\textsuperscript{~\orcidlink{0000-0001-9354-3909}}, Jiannong Cao\textsuperscript{~\orcidlink{0000-0002-2725-2529}}~\IEEEmembership{Fellow,~IEEE} 
\IEEEcompsocitemizethanks{
\IEEEcompsocthanksitem Tong Nie and Wei Ma are with the Department of Civil and Environmental Engineering, The Hong Kong Polytechnic University, Hong Kong, SAR, China (E-mail: tong.nie@connect.polyu.hk, wei.w.ma@polyu.edu.hk).
\IEEEcompsocthanksitem Jian Sun is with the Department of Traffic Engineering, Tongji University, Shanghai, China (E-mail: sunjian@tongji.edu.cn).
\IEEEcompsocthanksitem Yu Yang is with the Centre for Learning, Teaching, and Technology, The Education University of Hong Kong, Hong Kong, SAR, China (E-mail: yangyy@eduhk.hk).
\IEEEcompsocthanksitem Jiannong Cao is with the Research Institute for Artificial Intelligence of Things, Department of Computing, The Hong Kong Polytechnic University (E-mail: jiannong.cao@polyu.edu.hk). 
\IEEEcompsocthanksitem Corresponding authors: Jian Sun and Wei Ma.
\protect\\

}
}

%
%


\markboth{Journal of \LaTeX\ Class Files,~Vol.~xx, No.~x, xxx~2024}{Shell \MakeLowercase{\textit{et al.}}: Bare Demo of IEEEtran.cls for Computer Society Journals}

%



\IEEEtitleabstractindextext{%
\begin{abstract}
Urban time series, such as mobility flows, energy consumption, and pollution records, encapsulate complex urban dynamics and structures. However, data collection in each city is impeded by technical challenges such as budget limitations and sensor failures, necessitating effective data imputation techniques that can enhance data quality and reliability. 
Existing imputation models, categorized into learning-based and analytics-based paradigms, grapple with the trade-off between capacity and generalizability. 
Collaborative learning to reconstruct data across multiple cities holds the promise of breaking this trade-off. Nevertheless, urban data's inherent irregularity and heterogeneity issues exacerbate challenges of knowledge sharing and collaboration across cities. To address these limitations, we propose a novel collaborative imputation paradigm leveraging meta-learned implicit neural representations (INRs). INRs offer a continuous mapping from domain coordinates to target values, integrating the strengths of both paradigms. 
By imposing embedding theory, we first employ continuous parameterization to handle irregularity and reconstruct the dynamical system. We then introduce a cross-city collaborative learning scheme through model-agnostic meta learning, incorporating hierarchical modulation and normalization techniques to accommodate multiscale representations and reduce variance in response to heterogeneity. Extensive experiments on a diverse urban dataset from 20 global cities demonstrate our model's superior imputation performance and generalizability, underscoring the effectiveness of collaborative imputation in resource-constrained settings. 
\end{abstract}

}

\maketitle

\IEEEdisplaynontitleabstractindextext

%
\IEEEpeerreviewmaketitle

\section{Introduction}

\IEEEPARstart{T}{ime} series measured in urban agglomerations, such as mobility flows, energy consumption, and pollution records, represent time-dependent urban patterns and dynamics. These high-granular quantities can be exploited by data-driven models to reflect latent profiles of cities, such as human activity, socio-economic and welfare \cite{fan2023urban}. 
To utilize such data, either Eulerian sensors with dense spatial deployments or Lagrangian sensors with high temporal coverage are desired to measure them \cite{zheng2014urban}.
However, access to city-scale holographic data is far from easy due to factors such as installation and maintenance costs, adverse observation conditions, and system errors, hindering the usage of urban computing applications \cite{wang2020deep,jin2023spatio}. Therefore, data imputation technique has emerged to compensate for the lack of full observations and enhance data quality and reliability.

\begin{figure}[!htbp]
  \centering
  \captionsetup{skip=1pt}
  \includegraphics[width=1\columnwidth]{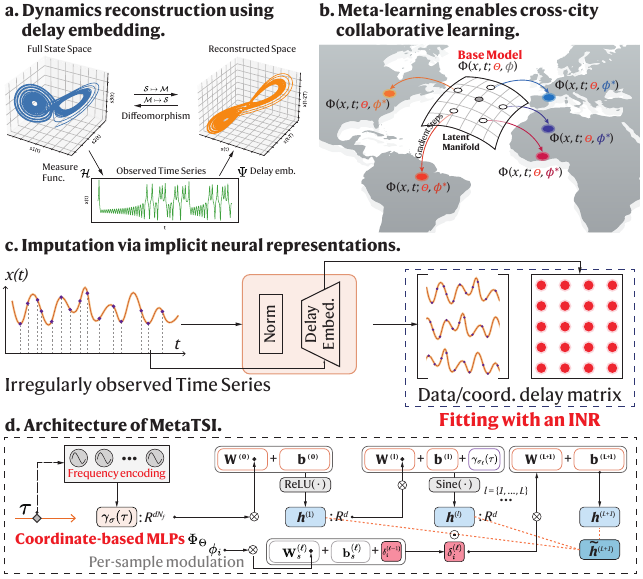}
  \caption{Cross-city collaborative imputation framework.}
  \label{fig:intro}
\end{figure}

Existing data imputation models generally fall into two paradigms: learning-based and analytics-based solutions. However, both paradigms face \textit{the dilemma of capacity and generalizability}. The former applies neural architectures that excel in fitting the time series distribution, such as RNNs \cite{cao2018brits,che2018recurrent}, probabilistic diffusion \cite{liu2023pristi,senane2024self}, and Transformers \cite{du2023saits,nie2024imputeformer}.
However, deep time series models struggle to comprehend underlying physical laws of data, leading to the potential for overfitting in masked training and lacking generalizability to unseen data outside of the training distribution \cite{han2024capacity}.
The latter formalism designs analytical models to characterize commonly shared data properties, showing better generalizability across data. One prominent example is the matrix completion model \cite{yu2016temporal,chen2024laplacian,asif2016matrix}. Unfortunately, they have limited model capacity and are restricted to fixed dimensions with a predefined input domain. 
As a result, they usually require per sample optimization, preventing them from modeling complex and diverse urban datasets.

Promisingly, the recent success of foundation models has demonstrated the potential of learning from the union of diverse datasets \cite{jin2023large,ma2024survey}. Large models pretrained on cross-source data hold promise for breaking the trade-off.
This inspires us to develop an expressive and generalizable imputation paradigm across multiple cities. However, public agencies in each city currently have to develop individual models based on separate expertise, which is resource-intensive with a specialized computational procedure. More critically, urban data within each city can only be used for the city and considered in isolation from each other. This hinders its accessibility and information share in resource-constrained contexts, where less developed cities have low observation rates and limited technical capacity to fully train a model. With a generalizable model and collaborative data governance, the shared knowledge from data-rich cities can inform different-but-related patterns in data-lacking cities.

Overall, the benefits of cross-city learning prompt us to explore the possibility of an innovative and alternative scheme for collaborative imputation.
To this end, we summarize the challenges in urban time series imputation under practical constraints of observation conditions as: \textit{irregularity and heterogeneity}. 
First, urban time series is irregularly sampled in nature.
The data might be generated as a burst or with varying time intervals, and different sensors can have different sampling frequencies. Furthermore, missingness can manifest itself at arbitrary locations and timestamps \cite{ma2019missing}.
To leverage state-of-the-art imputation models, one may convert an irregular time series into a regular time series. This can cause information loss and misinterpretation of the missing pattern.
Second, urban data displays high heterogeneity \cite{xue2022quantifying,Shao2024exploring}. Measurements from different locations show diverse localized patterns such as multiple scales and frequencies.
Consequently, collaborative learning from the joint distribution of heterogeneous data requires rigorous model design and large capacity. In addition, such heterogeneity increases the difficulty in discovering shared patterns that are generalizable between different cities.

In summary, most state-of-the-art models either apply instance-specific optimization with fixed spatial-temporal dimensions \cite{chen2024laplacian} or are trained to interpolate discrete signals in regular grids for a single data source \cite{nie2024imputeformer}. 
They are biased toward particular data sources and compromise between accuracy and generality depending on the applied domain \cite{du2021learning}, hindering knowledge transfer across heterogeneous cities.
To resolve this dilemma, we resort to an emerging approach called implicit neural representations (INRs). INRs have recently been shown to be proficient in learning representations from multimodal data, such as 2D images, 3D scenes, audios, time series, and spatiotemporal data \cite{SIREN,FourierFeature,mildenhall2021nerf,chen2021learning,luo2023low,woo2023learning,nie2024spatiotemporal}.
INRs represent data instances by parameters of neural networks and establish continuous mappings from domain coordinate to quantity of interests. They inherit merits of both two paradigms with high expressivity and universal priors such as smoothness.

By capitalizing on the recent advancements in INRs, 
we first tackle the irregularity using continuous parameterization. Then, under the embedding theory \cite{whitney1936differentiable,takens2006detecting}, the incomplete time series is imputed by reconstructing the dynamical system with global diffeomorphic embedding and a frequency-decomposed architecture.
To achieve collaborative modeling of commonalities and shared patterns across cities to remove data barriers, we frame it as a generalizable representation learning task and convert the data of cities to centralized weights of neural networks and distribute them to different cities to complete imputation tasks. This process is accomplished by a model-agnostic meta learning (MAML) \cite{finn2017model} framework in which an underlying low-dimensional manifold is discovered to learn the distribution of functions in a sensor-agnostic manner.
To further deal with heterogeneity in cross-city transfer, masked instance normalization is proposed to alleviate the difficulty of learning with city-specific variance, and a hierarchical and multiscale modulation mechanism is developed to condition the model on heterogeneous patterns in a parameter-efficient way.

Integrating both physics and data priors into a holistic meta-learned INR architecture, we develop a novel cross-city collaborative imputation scheme that preserves both expressivity and generalizability.
It enables the capture of shared knowledge across locations and efficient adaptation to new cities with few observations.
To evaluate it, we collect a large-scale urban dataset from 20 cities around the globe, consisting of more than 8,000 time series with different resolutions, sampling frequencies, and spatiotemporal patterns. Experimental results suggest that our model contributes to the state-of-the-art with better imputation accuracy and generalizability.
Our contributions are summarized as follows:

\begin{itemize}
\item We introduce a new time series imputation paradigm through INRs. By connecting the embedding theory with INRs, it reconstructs irregular urban time series with a frequency-decomposed deep architecture;
\item A novel cross-city collaborative learning scheme is proposed via MAML, featuring a hierarchical modulation mechanism for multiscale conditioning and normalization to reduce instance-level heterogeneity;
\item Empirical results show that our model outperforms SOTA baselines in both single-city and cross-city scenarios. The significance of the collaborative imputation scheme is also highlighted in few-shot settings.
\end{itemize}

The remainder is structured as follows. Sections \ref{sec:related_work} and \ref{sec:preliminary} introduce the related work and background information. Section \ref{sec: methodology} introduces the collaborative imputation framework. Section \ref{sec: analysis} provides algorithmic analysis to interpret the model.
Section \ref{sec:experiments} conducts experiments on a large-scale urban dataset. Finally, we conclude this study in Section \ref{sec: conclusion}.


\section{Related Work}\label{sec:related_work}

\noindent\textbf{Spatiotemporal data imputation.}
As missing data is pervasive in spatiotemporal system, many recent studies have been devoted to establishing data imputation models \cite{miao2022experimental}. Existing methods can be categorized into two types, i.e., learning-based and analytics-based methods. The former adopt deep neural networks to correlate observed values and learn to utilize the correlations to fill unobserved values. Popular architectures such as recurrent neural networks, diffusion models, and Transformers are widely adopted \cite{cao2018brits,che2018recurrent,du2023saits,liu2023pristi,nie2024imputeformer,senane2024self,ma2024steinformer,liu2024scope}.
Analytical models approach the data imputation problem by solving an optimization problem associated with missing data patterns \cite{karmitsa2020missing,gong2021missing,nie2022truncated,blazquez2023selective,luo2023low,wang2023low,chen2024laplacian}. Representative methods include low-rank matrix factorization and tensor completion.
The two existing paradigms facilitate the development of a specific imputation model for a particular dataset. Collaborative imputation of data across multiple datasets remains unexplored. In addition, although there are some imputation models designed for irregular time series \cite{shukla2018interpolationprediction,li2020learning,weerakody2021review,naour2023time}, their abilities in large-scale urban datasets are less explored.

\noindent\textbf{Implicit neural representations.}
INRs are a class of methods to implicitly define a quantity of interest by associating the target value with its coordinate. By fitting the given data using coordinate-based neural networks, INRs provide an alternative to store and query data. Due to their continuous, expressive, efficient properties, INRs have achieved success in learning multimodal data, such as images, videos, 3D scenes, point cloud, and audio data \cite{SIREN,FourierFeature,mildenhall2021nerf,chen2021learning,dupont2021coin,luo2023low}.
In addition, INRs have recently been adopted for spatiotemporal data \cite{luo2024continuous,nie2024spatiotemporal} and time series \cite{fons2022hypertime,naour2023time,woo2023learning}, with a main focus on reconstructing and forecasting tasks.
Most of these studies either concentrate on a single task from a particular data source or evaluate the performances of INRs on various different tasks in parallel.

\noindent\textbf{Meta-learning INRs.} As an INR is fitted to a single instance using weights of neural networks, the learned weights cannot be applied to predict other instances directly. To address this limitation, generalizable implicit neural representations (GINRs) are developed using gradient- or Transformer-based meta-learning algorithms \cite{nie2024generalizable,kim2023generalizable,functa,sitzmann2020metasdf,mehta2021modulated}. The mechanism of GINRs is to model the distribution of functional representations using separate parameters, making them adaptable to different instances with a shared meta model and specific task models. Prior research on GINRs has focused on training them on homogeneous data, such as images, with the objective of developing a task-independent decoder. The potential of GINRs for cross-city learning of large-scale urban datasets has yet to be fully investigated.

\noindent\textbf{Cross-city generalization.} To encourage data and resource sharing between cities, previous work develops transfer-learning schemes for cross-city learning, with a particular focus on traffic forecasting problems \cite{wang2019cross,tang2022domain,ouyang2023domain,jin2023transferable,ouyang2023citytrans,zhang2024personalized,wang2024cola}. With abundant observations available, the knowledge from source cities can be readily transferred to the target city. However, for the studied imputation task, cross-city generalization is more challenging with sparse data. In addition, as discrete neural networks are adopted as backbones, they can only deal with urban data with regular data organization, e.g., fixed spatial-temporal dimension.

\section{Preliminary}\label{sec:preliminary}
\noindent\textbf{{Notations}}. We denote matrices by boldface capital letters e.g., $\mX\in\mathbb{R}^{N\times T}$, vectors are denoted by boldface lowercase letters, e.g., $\vx\in\mathbb{R}^{T}$, and scalars are lowercase letters, e.g., $x$. 
Without ambiguity, we also denote $\vx(t)\in\mathbb{R}^m$ an arbitrary time series with time index $t$.
The functional representation (or a mapping function) is abbreviated as $\Phi(\cdot)$. 
In the absence of remarks, calligraphic letters are used to denote the vector space, for example $\mathcal{X}\subseteq\mathbb{R}$. 

\noindent\textbf{Model-agnostic meta-learning.} 
To utilize the shared pattern and knowledge between source and target data, a transfer learning algorithm needs to be established. The meta learning framework is developed for learning the learning algorithm itself. In particular, the model-agnostic meta-learning (MAML) algorithm \cite{finn2017model} is designed to train a deep neural network on a variety of learning tasks, such that the source model is explicitly fine-tuned to generalize to a downstream task with few labels using a small number of gradient steps.
MAML obeys the following iterative scheme:
\begin{equation}\label{eq:meta-learning}
\begin{aligned}
    &\underbrace{\min_{\Theta}\mathbb{E}_{\tau\sim p(\tau)}\left[\ell_{\tau}^{\text{test}}(\phi_{\tau,K}(\Theta))\right]}_{\text{outer loop/learning meta model}}, \\
    &\text{s.t.}~\underbrace{\phi_{\tau,k+1}\leftarrow\phi_{\tau,k}-\alpha\nabla_{\phi_{\tau,k}}\ell_{\tau}^{\text{train}}}_{\text{inner loop/learning specific model}},~\underbrace{\phi_{\tau,0}=\Theta}_{\text{initialization}},
\end{aligned}
\end{equation}
where a task-specific model $\phi_{\tau}$ on task $\tau$ is initialized by the parameter of meta-model $\Theta$ and iteratively updated with $K$ steps based on a supervision loss $\ell_{\tau}^{\text{train}}$ of a few training samples in the inner loop. In the outer loop, the meta parameters $\Theta$ are updated by minimizing the test loss $\ell_{\tau}^{\text{test}}$ over a batch of tasks $\mathbb{E}_{\tau\sim p(\tau)}$ with the gradient-adapted parameters $\phi_{\tau,k}$. 
MAML yields a meta-model that is explicitly fine-tuned to differentiate different instances.

\noindent\textbf{{Problem formulation}}. If all time series are recorded synchronously and the sensor number is fixed during the observation period, the data can be organized as a spatiotemporal matrix $\mX\in\mathbb{R}^{N\times T}$ with $\vx^{i}\in\mathbb{R}^{T}$ denoting the $i$-th series and $\vx_{t}\in\mathbb{R}^{N}$ indicating observations at time $t$. Then the urban data imputation problem can be considered as estimating a probability $\forall \tau\in\{1,\dots,T\},n\in\{i=1,\dots,N\}$:
\begin{equation}
    p(\{\vx_{\tau}^{n}|m_{\tau}^{n}=0\}|\{\vx_t^i|m_t^i=1\}_{t=1,\dots,T}^{i=1,\dots,N}),
\end{equation}
where $m_t^i\in\{0,1\}$ is an indicator of observable which is 1 if the measurements associated with the $i$-th sensor are valid at time step $t$. This posterior probability is estimated by a parameterized model (such as a neural network) by:
\begin{equation}
    \Phi(\Omega(\mX)|\Theta)\approx\mathbb{E}[p(\{\vx_{\tau}^{n}|m_{\tau}^{n}=0\}|\{\vx_t^i|m_t^i=1\})],
\end{equation}
where $\Omega(x_t^i)=x_t^i$ if $m_t^i=1$.

In the case of irregularly sampled data, each time series sampled from different cities or sensors can have different sampling frequencies and dimensions. We use a sequence of data pairs to denote this irregularity as $\vx^i=\{(x_{t}^i,\xi^i,\tau_t)\}_{t=1}^{T_i}$ where the observation is paired with its location and time-index feature $(\xi^i,\tau_t)$. To mitigate the difficulty, we assume that the location of a sensor is fixed and the total number of sensors for a city remains unchanged during the period. Then the parameterized imputer is learned by minimizing the empirical loss over all time series in the training set:
\begin{equation}\label{eq:single_inr}
\begin{aligned}
    \min_{\Theta}~&\mathcal{L}(\{\vx^i\}_{i=1}^N;\Theta)=\mathbb{E}_{\vx\sim p_{\vx}}[\ell(\vx^{i};\Theta)], \\
   &\text{with~}\ell(\vx^{i};\Theta)=\frac{1}{T_i}\sum_{t=1}^{T_i}\Vert x_{t}^i-\Phi_{\Theta}(\xi^i,\tau_t) \Vert_2^2,
\end{aligned}
\end{equation}
where $p_{\vx}$ is the (unknown) data distribution. $\Phi$ is either a unified model or a composition of all individual models $\Phi=\{\Phi_1,\cdots,\Phi_N$\}, which corresponds to the \textbf{collaborative imputation} and \textbf{separate imputation} schemes respectively.
To deal with irregular data, a continuous model $\Phi$ is expected to query data at arbitrary location and timestamp.

\section{Methodology}\label{sec: methodology}
This section elaborates the proposed model for cross-city collaborative imputation.
We aim to learn an INR $\Phi_{\Theta}(\xi,\tau):\mathbb{R}\times\mathbb{R}\mapsto\mathbb{R}$ that maps the timestamp (possibly with location) to the observed value $x$ at each query coordinate. Then the continuous function $\Phi_{\Theta}(\cdot)$ allows for interpolation at arbitrary points, generating the imputation for irregular time series. The meta-learning method further enables it to generalize across heterogeneous cities.
We term it \underline{Meta} learning-based urban \underline{T}ime \underline{S}eries \underline{I}mputer (MetaTSI).
The overall architecture of MetaTSI is shown in Figure \ref{fig:intro} (c).

\subsection{Learning to Reconstruct Time Series Dynamics by Implicit Representation in the Embedding Space}

\subsubsection{Implicit Neural Representations for Time Series}
Urban system involves complex spatiotemporal phenomena. The sensed time series may be generically generated from a high-dimensional dynamical system \cite{whitney1936differentiable}:
\begin{equation}\label{eq:inr}
    \mathcal{C}(t,\vx(t),\nabla_{t}\vx(t),\nabla^2_{t}\vx(t),\dots)=0,
\end{equation}
where the time series $\vx(t):t\mapsto\mathbb{R}^m$ from a data generating process consists of successive, possibly irregular, observations of some dynamical process described by a system of partial differential equation $\mathcal{C}$.
In this context, our goal is to learn a parameterized neural network $\Phi$ to map $t$ to the target value $\vx(t)$ while reconstructing the dynamics in Eq. \ref{eq:inr}, achieving the imputation in observations:
\begin{equation}
    \min \mathcal{L} = \int_{\mathcal{T}}\Vert\mathcal{C}(\vx(t),\Phi,\nabla_{t}\Phi,\nabla^2_{t}\Phi,\dots)\Vert dt,
\label{eq:inr_intro}
\end{equation}
where $\mathcal{T}$ is the definition domain and the loss is feasible by sampling a dataset $\mathcal{D}=\{(\vx(t),t):t\in\mathcal{T}\}$ of coordinates and observations.
Since $\Phi$ is implicitly defined by the relation modeled by $\mathcal{C}$, neural networks that parameterize such implicitly defined functions are referred to as INRs \cite{SIREN}. 
Typically, fitting an INR requires a large amount of observed data \cite{SIREN}. Directly learning INRs from sparse time series can be challenging and suboptimal due to the lack of physical guidance and data properties. To address this issue, we exploit methods that can guide the learning process.

\subsubsection{Dynamics Reconstruction by the Embedding Theory}

To obtain such a model $\Phi$, we resort to the embedding theory \cite{whitney1936differentiable,takens2006detecting}. Generally, the dynamical system can be characterized by a state variable $\vs(t)\in\mathbb{R}^n$ and a measurement function $\mathcal{H}: R^n \mapsto R^m$. Typically, $m<n$ as the state of the underlying dynamical process cannot be fully observed. For urban time series, we will consider in this paper the simplest case where $x(t)$ is a scalar (i.e. $m = 1$) time series. Given a partially observed system state $\vs(t)$ with dynamics on state space $\mathcal{S}\subseteq\mathbb{R}^n$, it can be reconstructed by:
\begin{equation}
    \vec{\vs}(t)=f(\vs(t)),
\end{equation}
where $f(\cdot)$ is the reconstructor defined in the state space.

The measurement function $\mathcal{H}$ specifies the process of observing the system and extracting information evaluated at query time steps to generate a time-series:
\begin{equation}
    x(t)=\mathcal{H}({\vs(t)}).
\end{equation}

Then, an embedding is defined as a transformation that augments the observed time series by increasing its dimension with some time windows:
\begin{equation}
    \Psi:\mathcal{R}\mapsto\mathcal{R}^{d_r}, ~\vec{\vx}(t)=\Psi(x(t)),
\end{equation}
where $\vec{\vx}(t)$ is the embedding vector with dynamics defined in a reconstructed state space $\vec{\mathcal{M}}\subseteq R^{d_r}$ and a transformed reconstructor $\mathcal{F}$.
According to Takens' theory \cite{takens2006detecting}, if $\Psi$ is a valid embedding, there exists a one-to-one mapping $\Psi$ that we can identify a corresponding state point $x(t)$ on $\mathcal{S}$ for every $\vec{\vx}(t)$ on $\vec{\mathcal{M}}$ via the inverse mapping $\Psi^{-1}(\vec{\vx}(t))$:
\begin{equation}
    \Psi^{-1}:\vec{\mathcal{M}}\mapsto\mathcal{S},\Psi^{-1}(\vec{\vx}(t))=x(t),
\end{equation}
and the the dynamics of the system are preserved by:
\begin{equation}
    \mathcal{F}=\Psi\circ f\circ\Psi^{-1}.
\end{equation}

We can find that learning the dynamics in the reconstructed state space $\vec{\mathcal{M}}$ is equivalent to learning the dynamics of the original system $\mathcal{S}$. Therefore, the time series imputation problem using embedding is posed as learning the parameterized reconstructor $\mathcal{F}_{\Theta}$ where:
\begin{equation}\label{eq:reconstructor}
    \tilde{x}(t)=\Psi^{-1}\circ\mathcal{F}_{\Theta}\circ\Psi(x(t)).
\end{equation}

As described in Eq. \ref{eq:reconstructor}, the key to modeling the underlying dynamical system for imputation is to build a suitable data embedding $\Psi$ and a mapping $\mathcal{F}$.
Rather than relying on neural networks to directly process the raw data and correlate hidden states by temporal modules, we reconstruct the dynamical trajectory of the system using a nonparametric physical embedding as $\Psi$ and parameterized INRs as $\mathcal{F}_{\Theta}$. We will elaborate on each of them in the following.

\begin{remark}[Relationship between data-driven time series modeling and implicit time series representation.]
There are two principal ways to model time series. Data-driven time series models directly embed time series into hidden space using trainable transformation, relying on neural networks to model data correlations based on hidden representations. However, INRs embed the spatial-temporal coordinate in the state space using nonparametric techniques and reconstruct the dynamical structure, where observations are used as labels rather than input.
According to Embedding Theory \cite{sauer1991embedology,takens2006detecting}, dynamical systems and time series can be mutually transformed through observation functions and reconstruction operation. 
\end{remark}

\subsubsection{Coordinate Delay Embedding}

According to Takens’ embedding theorem,
it is guaranteed that a time delay embedding \cite{packard1980geometry} with dynamics defined in a space of sufficiently large dimension $\mathbb{R}^{d_r}$ constructed from scalar time series is generically diffeomorphic to the full state space dynamics of the underlying system \cite{wu2024predicting}. 
This facilitates the recovery of full-state dynamics using partially observed time series.
To achieve this, time delay embedding augments a single scalar time series $x(t)$ to a higher dimension by constructing a delay vector $\vec{\vx}(t)\in\mathbb{R}^{m\delta}$:
\begin{equation}
    \vec{\vx}(t) = \Psi(x(t)) =[x(t),x(t+\delta),\dots,x(t+(m-1)\delta)]^{\mathsf{T}},
\end{equation}
where $\delta$ is the delay lag, $m$ is the embedding dimension, and $d_r=m\delta$.
Considering a given window $T$, the embedding matrix $\vec{\mX}=\Psi(\vx)\in\mathbb{R}^{(T-(m-1)\delta)\times m\delta}$ is given by:
\begin{equation}\label{eq:delay_emb}
   \vec{\mX}=\Psi(\vx) =  [\Psi(x(1)),\Psi(x(2)),\dots,\Psi(x(T-(m-1)\delta))]^{\mathsf{T}},
\end{equation}
which is expanded as:
\begin{equation}
\begin{bmatrix}
x_1 & x_{1+\delta} & \cdots & x_{1+(m-1)\delta} \\
x_2 & x_{2+\delta} & \cdots & x_{2+(m-1)\delta} \\
x_3 & x_{3+\delta} & \cdots & x_{3+(m-1)\delta} \\
\vdots & \vdots & \ddots & \vdots \\
x_{T-(m-1)\delta} & x_{T-(m-2)\delta} & \cdots & x_{T}
\end{bmatrix}.
\end{equation}

For a multivariate system with $N$ components, one can reconstruct a topologically isomorphic manifold $\vec{\mathcal{M}}^i$ from every series $\vx^i$ within the system $(i=1,\dots,N)$ in a $d_r$-dimensional space. This separate treatment aligns with the channel independence strategy for time series \cite{han2024capacity,nie2024channel}.

Recall that the input to our model $\Phi$ is the coordinate in the reconstructed state space $\vec{\mathcal{M}}$. Different from the standard time delay embedding, we propose structuring the domain coordinate with delay embedding, which leads to the following relation in the state space:
\begin{equation}
    \hat{\Psi}(\vx) \approx \mathcal{F}\circ\Psi(\bm{\tau})\in\mathbb{R}^{(T-(m-1)\delta)\times m\delta},
\end{equation}
where $\bm{\tau}=[\tau_1,\tau_2,\dots,\tau_T]^{\mathsf{T}}\in\mathcal{T}$ is the vector of time-index feature. $\Psi$ organizes the index into a structured coordinate-delayed matrix, and $\mathcal{F}$ maps it to state values in higher dimensions.
An inverse transform is adopted to embed the reconstruction to the original space through Eq. \ref{eq:reconstructor} as:
\begin{equation}\label{eq:delay_emb_reverse}
    \Psi^{-1}\circ\hat{\Psi}(\vx) = \mD^{\dag}\operatorname{vec}(\hat{\Psi}(\vx)) \in\mathbb{R}^T,
\end{equation}
where $\mD\in\{0,1\}^{(T-(m-1)\delta)\times m\delta}$ is a duplication matrix and $\mD^{\dag}=(\mD^{\mathsf{T}}\mD)^{-1}\mD^{\mathsf{T}}$ is the pseudo-inverse of $\mD$.

Our approach prescribes the global structure of the reconstructed space:
$\vec{\mathcal{M}}^i$ also has an isomorphic topological structure with the original system and the reconstruction preserves the properties of the dynamical system that do not change under smooth coordinate changes, ensuring a continuous imputation function.

\begin{remark}[Relationship with deep time embedding]
Many deep time embedding methods such as RNNs, casual attention, and reservoir computing, may also be related to delay embedding. 
In these models, the input time series is fed into a parameterized network that directly projects it into hidden states. The forward propagation of past states on current states effectively acts as a time delay embedding with small delay lag and large embedding dimension.
Instead, our model reconstructs the dynamical trajectory of the system using physical priors as data embedding.
\end{remark}



\subsubsection{Frequency-decomposed Multi-scale INRs}\label{sec:architecture}

Next, we detail the architecture and parameterization of the reconstructor $\mathcal{F}_{\Theta}\overset{\cdot}{=}\Phi_{\Theta}$. It consists of two components: a mapping network and a modulation network. The former maps coordinates from the state space to target values in that location. The latter is used to adapt the model to different instances, which will be discussed in Section \ref{sec:meta-learning}.
Adhere to previous practices \cite{luo2023low}, we adopt a factorized functional representation to reduce the difficulty and complexity of learning in the entire space. By denoting the transformed coordinate in the embedding space as $\bm{\vec{\tau}}_t$, we have:
\begin{equation}\label{eq:factorize}
\begin{aligned}
    &\Phi_{\Theta}(\bm{\vec{\tau}}_t) = \mathcal{C}\times_1\Phi_{\theta^1}(\vec{\tau}_{t}^1)\times_2\Phi_{\theta^2}(\vec{\tau}_{t}^2), \forall (\vec{\tau}_{t}^1,\vec{\tau}_{t}^2)\in\bm{\vec{\tau}}_t,\\
    &\Phi_{\theta^i}: \vec{\tau}_{t}^i\mapsto \Phi_{\theta^i}(\vec{\tau}_{t}^i)\in\mathbb{R}^{n_i},~\forall i\in\{1,2\},
\end{aligned}
\end{equation}
where $\mathcal{C}\in\mathbb{R}^{n_1\times  n_2}$ is the core tensor, $\times_i$ is the $i$-th mode tensor product, $\vec{\tau}_{t}^i$ is the embedded coordinate in the $i$-th axis,
and $\Theta=\{\theta^1,\theta^2\}\cup\mathcal{C}$ are trainable parameters. 
Each $\Phi_{\theta^i}$ can be defined separately to consider different patterns in the state space, and the interaction between the two components is preserved by $\mathcal{C}$. In practice, deep neural networks can become parameterization. However, learning to regress the target value in a high-dimensional system using the state coordinate is an ill-posed problem, as the associated neural tangent kernel is nonstationary \cite{FourierFeature}, which is understood as the ``spectral bias'' issue.

There are several ways to alleviate this bias, including periodic activations \cite{SIREN}, Fourier features \cite{FourierFeature}, polynomial functions \cite{singh2023polynomial}, and multiplicative filters \cite{fathony2020multiplicative}.
We consider a multilayer and multiscale structure to exploit the complex spatiotemporal structure of urban data. Given $\forall\ell=\{1,\dots,L\}$, each of the mapping subnetwork $\Phi_{\theta}(\vec{\tau}_t^i): \vec{\tau}_t^i\in\mathbb{R}\mapsto\Phi_{\theta}(\vec{\tau}_t^i)\in\mathbb{R}^{n_i}$
is formulated as:
\begin{equation}\label{eq:freqmlp}
\begin{aligned}
    &\vh^{(1)} = \texttt{ReLU}(\mathbf{W}^{(0)}\gamma_{\sigma}(\vec{\tau}_t^i)+\vb^{(0)}), \\
    &\vh^{(\ell+1)} = \delta^{(\ell)}_i \odot \sin(\mathbf{W}^{(\ell)}\vh^{(\ell)}+\vb^{(\ell)}+\gamma_{\sigma_{\ell}}(\vec{\tau}_t^i)),\\
    &\tilde{\vh}^{(L+1)} = \mathbf{W}^{(L+1)}\vh^{(L+1)}+\vb^{(L+1)},
\end{aligned}
\end{equation}
where $\mathbf{W}^{(\ell)}\in\mathbb{R}^{d_{(\ell+1)}\times d_{(\ell)}},\mathbf{b}^{(\ell)}\in\mathbb{R}^{d_{(\ell+1)}}$ are layerwise parameters with $d_{(0)}=d_B{N_f}$ being the input dimension and $d_{(L+1)}=n_i$ being the output dimension, $\delta^{(\ell)}_i\in\mathbb{R}^{d_{(\ell+1)}}$ is the modulation variable in the $\ell$-th layer described in Section \ref{sec:modulation}, and $\odot$ is the element-wise product.
$\gamma_{\sigma}(\cdot)$ is the concatenated Fourier features (CRF) with basis frequency $\mathbf{B}_k\in\mathbb{R}^{d_B/2\times 1}$ sampled from a Gaussian $\mathcal{N}(0,\sigma_{k}^2)$:
\begin{equation}\label{eq:crf}
\begin{aligned}
    \gamma(c)_{\sigma}=[&\sin(2\pi\mathbf{B}_1c),\cos(2\pi\mathbf{B}_1c),\dots, \\
    &\sin(2\pi\mathbf{B}_{N_f}c),\cos(2\pi\mathbf{B}_{N_f}c)]^{\mathsf{T}}\in\mathbb{R}^{d_B{N_f}}.
\end{aligned}
\end{equation}

There are several key points in Eq. \ref{eq:freqmlp} that need to be emphasized. (1) Both the sine activation and CRF are adopted to explicitly inject high-frequency structures into the network. (2) In particular, $\gamma_{\sigma_{\ell}}(\vec{\tau}_t^i)$ is the Fourier feature in the $\ell$-th layer. Although CRF in the input can reduce the spectral bias, a simple stack of MLPs still suffers from capturing high-frequency data details \cite{lee2024locality,ma2024tinyvim}. Instead, we decompose intermediate features into multiple frequency features to amplify high-frequency patterns by using different spectral bandwidths in different layers: $\sigma_1\geq \sigma_2\geq\dots\sigma_{L}$. Then the compositional nonlinearity of deep neural networks is applied recursively to progressively decode the multi-band intermediate features, achieving its representational complexity to model high-frequency patterns in deeper layers.

Consequently, a layerwise output is constructed that assembles the final reconstruction of $\Phi_{\theta^i}$:
\begin{equation}
    \tilde{\vh}^{(\ell)} = \texttt{ReLU}(\mathbf{W}_{\text{out}}^{(\ell)}\vh^{(\ell)}+\vb_{\text{out}}^{(\ell))}), ~\Phi_{\theta}(\vec{\tau}_t^i) = \sum_{\ell=1}^{L+1}\tilde{\vh}^{(\ell)}.
\end{equation}

Residual connections of all intermediate features into the output synthesize the multiscale reconstruction of the dynamical space and effectively predict details of data.
Reconstruction from low-level to high-level frequency features resembles the coarse-to-fine approach in the spatial domain.

\subsection{Cross-city Generalization by Meta Learning}\label{sec:meta-learning}

We remark that an INR can encode a single time series by fitting to observed values. However, adopting individual INRs to memorize each instance in each city confronts two problems: (1) lack of sufficient observations in less developed cities poses challenges in model training; (2) having individual models for each location is computationally expensive and infeasible.
A collaborative imputation scheme for learning across cities with limited resources that can be reused in new cities is preferable.
However, learning across cities creates a consequent heterogeneity issue.
Therefore, we resort to the meta learning scheme that allows us to learn the manifold in which the signals reside in a sensor-agnostic way, over arbitrary measurement from different locations. We further handle heterogeneity by proposing a normalization method and a modulation mechanism to condition the model to differentiate heterogeneous individual patterns.


\subsubsection{Masked Instance Normalization}
Urban time series feature large individual-level variations.
Urban networks themselves are heterogeneous around the world \cite{xue2022quantifying}. Furthermore, mobility patterns within each city vary by space and time, causing the sensed urban time series to fluctuate drastically.
To reduce individual-level variance for better generalization, we propose a masked instance normalization strategy. Given the observed time series $\vx^{i}\in\mathbb{R}^{T_i}$, we normalize it before the delay embedding:
\begin{equation}\label{eq:normalization}
    \begin{aligned}
        &\mathbb{E}_t[\vx^{i}]=\frac{1}{|\{m_{t}^i\}_{t=1}^{T_i}|}\sum_{t=1}^{T_i}m_{t}^i\odot x_{t}^i, \\
        &\text{Var}[\vx^{i}]=\frac{\sum_{t=1}^{T_i}(x_{t}^i-\mathbb{E}_t[\vx^{i}])^2}{|\{m_{t}^i\}_{t=1}^{T_i}|}, \hat{x}_{t}^i= \frac{x_{t}^i-\mathbb{E}_t[\vx^{i}]}{\sqrt{\text{Var}[\vx^{i}]+\epsilon}},
    \end{aligned}
\end{equation}
where $m_{t}^i$ is the masking indicator defined in Section \ref{sec:preliminary}, $\hat{x}_{\tau}^i$ is the normalized labels used for supervision. Normalized sequences can have a more consistent mean and variance, where the distribution discrepancy between different instances is reduced. This makes it easier for the model to learn local dynamics within the sequence while receiving input of consistent distributions.
However, the input has statistics different from the original distribution. 
A de-normalization step is needed to inform the model the original distribution of the instance by returning the distribution properties removed from the input to the model output:
\begin{equation}\label{eq: denormalization}
    \hat{\Phi}_{\Theta}(\tau_{t}) = \sqrt{\text{Var}[\vx^{i}]+\epsilon} \cdot \Phi_{\Theta}(\tau_{t}) + \mathbb{E}_t[\vx^{i}].
\end{equation}

\subsubsection{Meta-learning-based Model Generalization}

As stated above, a coordinate-based MLP (i.e., INR) can learn to represent each data instance, but the learned MLP cannot be generalized to represent other instances and requires re-optimizing from the scratch.
Given a set of $N$ data instances $\mathcal{X}=\{\vx^{i}\}_{i=1}^N$ with $\vx^i=\{(x_{t}^i,\tau_t)\}_{t=1}^{T_i}$, a simple approach to represent the entire dataset is to train an individual INR $\Phi_{\theta^i}$ for each instance ($\mathcal{R}(\cdot)$ is omitted):
\begin{equation}
    \ell_i(\vx^{i};\Theta^i)=\frac{1}{T_i}\sum_{t=1}^{T_i}\Vert x_{t}^i-\Phi_{\Theta^i}(\tau_t) \Vert_2^2,
\end{equation}
where $\mathcal{H}_{\text{INR}}=\{\Phi(\tau_t;\Theta^i)|\Theta^i\in\Theta,i=1,\dots,N\}$ is the hypothesis class of all INRs where $\Theta$ is the parameter space. However, such a hypothesis class is too expensive and computationally infeasible for a large number of instances in urban setting. In this case, standard INRs just memorize each data instance without generalization ability.

To enable INRs to be generalizable to different input instances in a memory-efficient way, we split the parameter space into two parts: (1) instance-specific parameters $\phi^i\in\Xi$, and (2) instance-agnostic parameters $\Theta$. $\phi^i$ characterizes each data instance and aims to learn to adapt to specific patterns. $\Theta$ is shared across all instances and designed to learn the inductive bias, e.g., the underlying structural information in urban time series. Then, the loss of generalizable implicit neural representations (GINRs) is given as:
\begin{equation}\label{eq:ST-INR_loss}
\begin{aligned}
    &\min_{\Theta,\phi}\ell(\mathcal{X};\Theta,\{\phi^{i}\}_{i=1}^N)=\mathbb{E}_{\vx\sim\mathcal{X}}[\ell^i(\vx^{i};\Theta,\phi^{i})], \\
    &=\frac{1}{N}\sum_{i=1}^N\frac{1}{T_i}\sum_{t=1}^{T_i}\Vert x_{t}^i- 
    \Phi_{\Theta,\phi}(\tau_t;\phi^{i}) \Vert_2^2, 
\end{aligned}
\end{equation} 
where $\phi_{i}\in\mathbb{R}^{d_{\text{latent}}}$ is the latent code for each data instance to account for the instance-specific data pattern.

$\Theta$ and $\phi^i$ characterize different perspectives of STTD and are nontrivial to obtain properly.
To consider the per-instance variations, a natural way is to encode $\phi$ from the observed data and map it to parameters of a base network, e.g., all linear weights in Eq. \ref{eq:freqmlp}. 
Consider the parameter that fully defines a $L$-layer single INR is $\boldsymbol{\theta}\in\mathbb{R}^D$ where $D=\sum_{\ell=1}^L d_{(\ell)}$, we can explicitly encode the observation $\vx^i\in\mathbb{R}^{T_i}$ to a low-dimensional vector $\mathbf{c}^i\in\mathbb{R}^C$, then decode it to the hypothesis class of $\Phi$ with a hypernetwork $h_{\text{hyper}}$:
\begin{equation}\label{eq:encoding}
    \begin{aligned}
        \texttt{Enc}&: \mathbb{R}^{T_i}\mapsto\mathbb{R}^C, \vx^i\mapsto\texttt{Enc}(\vx^i):=\mathbf{c}^i, \\
        h_{\text{hyper}}&: \mathbb{R}^C\mapsto\mathbb{R}^D, \mathbf{c}^i\mapsto h_{\text{hyper}}(\mathbf{c}^i)=\boldsymbol{\theta}, \\
    \end{aligned}
\end{equation}
where in this case $\phi=\{\mathbf{c}^i\}_{i=1}^N$ are the instance-specific parameters. However, manipulating the entire MLP can be parameter intensive and cause suboptimal representations. This process still admits a large parameter space as both the observation and hypernetwork can have a large dimension.

Instead, we treat $\phi^i$ as a learnable code and resort to the modulation method. We modulate the intermediate features of $\Phi_{\theta}$ per instance using the compact form of $\phi^i$. This is achieved by modifying only a few parameters of $\Phi_{\Theta}$ through a modulation network discussed in Section \ref{sec:modulation}.
The conditioning modulations are processed as a function of $\phi^i$, and each $\phi^i$ characterizes a specific instance. We can then \textit{implicitly} obtain these latent codes as well as modulations using an \textit{auto-decoding} mechanism \cite{park2019deepsdf}, instead of the explicit encoding process in Eq. \ref{eq:encoding}. 
We suppose that the internal manifold of correlated time series exists in a structured low-dimensional subspace that is globally consistent. Similar data samples should be embedded in a close location with small encoding steps. 
For data $i$, this is calculated by an iterative gradient descent process $\forall i=1,\dots, N$:
\begin{equation}\label{eq:auto-decoding}
    \phi^{(k+1),i}\leftarrow\phi^{(k),i}-\alpha\nabla_{\phi^{i}}\ell(\Phi_{\Theta,h_{\omega}{(\phi)}},\{\vx^i\}_{i\in \mathcal{B}}),
\end{equation}
where $\alpha$ is the learning rate, $h_{\omega}$ is the hypernetwork that generates modulations from the latent code to condition INRs,
and $\mathcal{B}$ is the sampled data batch. 
To initialize the learnable latent codes, a prior class over $p(\phi)$ is assumed to follow a zero-mean Gaussian.
With this, the reduced hypothesis class is $\mathcal{H}_{\text{MetaTSI}}=\{\Phi(\tau;\Theta,\phi^{i})|\Theta,\phi^{i}\in\mathbb{R}^{d_{\text{latent}}}\}$, which is more feasible in optimization.

The intuition of Eq. \ref{eq:auto-decoding} is to learn some priors over the function space of neural fields. However, learning all latent codes efficiently with a single base network is a challenge. Consequently, to integrate auto-decoding into the parameter learning procedure of the base network, a meta-learning scheme including inner loop and outer loop iterations is considered \cite{finn2017model}, as described in Eq. \ref{eq:meta-learning}. The aim of meta-learning is to learn the base parameter conditioning on the latent code, so that the code can be auto-decoded in a small number of iterations. Therefore, (i) the inner loop learns to adapt $\phi$ to condition the base network $\Phi^i$ on the instance-specific pattern $\vx^{i}$, and (ii) the outer loop learns to optimize the shared base parameters. 
In practice, the latent code is optimized by a small number of gradient descent steps in the inner loop (e.g., $N_{\text{inner}}=3$ is sufficient). Additionally, multivariate correlations are implicitly modeled by the interaction of latent code during optimization.

\subsubsection{Hierarchical Modulation Mechanism}\label{sec:modulation}
We now introduce the modulation network. Recall that the key of MAML is the fast adaptation to local variations with minimal modification of network weights. However, due to high heterogeneity of spatiotemporal patterns of urban data, manipulating only a single weight matrix is inadequate for complex datasets. To this end, we propose to adapt the model to different instances by modulating hierarchical hidden activations of the mapping subnetwork in Eq. \ref{eq:freqmlp}. Motivated by \cite{functa}, we model $\phi$ as a series of latent codes $\{\phi^{i}\in\mathbb{R}^{d_{\text{latent}}}\}_{i=1}^N$ for each instance to account for the instance-specific data pattern and make $\Phi_{\theta}$ a base network conditional on the latent code $\phi$. The per-sample modulations $\vs^{i}$ are considered as a function conditioned on the latent code $\phi^{i}$ that represents the individual time series:
\begin{equation}
\begin{aligned}
    \delta^{(\ell),i} &= \vs^{i}+\delta^{(\ell-1),i},~\ell=\{1,\dots,L\}, \\
    \vs^{i} &= h_{\omega}^{(\ell)}(\phi^{i})=\mathbf{W}^{(\ell)}_s\phi^{i}+\vb^{(\ell)}_s, \\
    \delta^{(0),i} &= \mathbf{W}^{(0)}_s\phi^{i}+\vb^{(0)}_s, 
\end{aligned}
\end{equation}
where $\delta^{(\ell),i}$ is the modulation of instance $i$ at layer $\ell$, and $h_{\omega}^{(\ell)}(\cdot|\omega\in\theta):\mathbb{R}^{d_{\text{latent}}}\mapsto\mathbb{R}^{d_{(\ell)}}$ is a shared linear hypernetwork to map the latent code to modulations, which enables one to modulate the weight space in a low-dimensional space, thereby significantly reducing the parameters. Note that the parameters $\{\mathbf{W}_s^{(\ell)},\mathbf{b}_s^{(\ell)}\}_{\ell=1}^L$ of $h_{\text{hyper}}$ is a subset of $\theta$. As can be seen, the modulation is established in a skip-connected structure. In addition, modulation vectors with high-frequency bandwidth can be processed with more nonlinear operations than vectors with lower frequencies, which hierarchically modulate complex signals.

In summary, we provide the workflow of MetaTSI that includes both the training stage and the inference stage in Algorithm \ref{algo:meta-learning}. During training, the knowledge from source cities is encoded in weights of deep neural networks and inform the fast adaptation of downstream cities.
During inference, the latent code $\phi$ adapts quickly to the current data pattern in just a few gradient steps. This mechanism makes MetaTSI capable of generalizing to new cities efficiently, which is a significant advantage over individual models.

\begin{algorithm}[!htb]
\setstretch{1}
\caption{MetaTSI for Cross-city Learning}\label{algo:meta-learning}
\begin{footnotesize}
\KwIn{Base network parameter $\Theta$, hypernetwork parameter $\omega$, training dataset $\mathcal{X}$.}
\KwOut{Trained base model $\Phi_{\Theta}$ and latent codes for all instances $\{\phi^{i}\}_{i=1}^N$.}
\tcp{Model training stage}
\While{not convergence}{
Sample a batch $\mathcal{B}$ of data $\{\{(x_{t}^i,\tau_t)\}_{t=1}^{T_i}\}_{i\in\mathcal{B}}$; \\
Normalize the observation $\hat{x}_{t}^i$ using Eq. \ref{eq:normalization}; \\
Embed $\{\{(\hat{x}_{t}^i,\tau_t)\}_{t=1}^{T_i}\}_{i\in\mathcal{B}}$ using Eq. \ref{eq:delay_emb}; \\
Set latent code to zeros $\phi^{i}\leftarrow 0,\forall i\in\mathcal{B}$;\\
\tcp{Inner loop for latent modulations}
\For{$s=1:N_{\text{inner}}$ and $i\in\mathcal{B}$}{$\phi^{i}\leftarrow\phi^{i}-\alpha\nabla_{\phi}\ell(\Phi_{\Theta,h_{\omega}{(\phi)}},\Psi(\{(\hat{x}_{t}^i,\tau_t)\}_{t=1}^{T_i}))|_{\phi=\phi^{i}}$;\\
}
\tcp{Outer loop for updating base parameters}
$[\Theta,\omega]\leftarrow [\Theta,\omega] - \eta \nabla_{\Theta,\omega}\frac{1}{|\mathcal{B}|}\sum_{j\in\mathcal{B}}\ell(\Phi_{\Theta,h_{\omega}{(\phi)}},\Psi(\{(\hat{x}_{t}^i,\tau_t)\}_{t=1}^{T_i}))|_{\phi=\phi^{j}}$;\\
Recover outputs to the original space using Eq. \ref{eq:delay_emb_reverse}; \\
De-normalize the reconstruction using Eq. \ref{eq: denormalization}; \\
}
\tcp{Model inference stage}
Given a partially observed new instance in grid $\mathcal{M}$ $\{\tau^{\star}_t,\vx_{t}^{\star}\}_{t\in \mathcal{M}}$, set $\phi^{\star}\leftarrow 0$ ;\\
Perform normalization and embedding; \\
\For{$s=1:N_{\text{inner}}$}{$\phi^{\star}\leftarrow\phi^{\star}-\alpha\nabla_{\phi}\ell(\Phi_{\theta,h_{\omega}{(\phi^{\star})}},\{\tau^{\star}_t,\vx_{t}^{\star}\}_{t\in \mathcal{M}}|_{\phi=\phi^{\star}}$;\\
}
Evaluate $\Phi_{\theta,h_{\omega}(\phi^{\star})}(\tau^{\star})$ for any $\tau^{\star}\in\mathcal{M}$.
\end{footnotesize}
\end{algorithm}

\subsection{Algorithmic Analysis}\label{sec: analysis}
This section provides further discussion on the architectural bias of MetaTSI. 
We theoretically demonstrate that the mapping network is a continuous function of the target signal and can be viewed as a composition of Fourier series.

\subsubsection{MetaTSI as Continuous Functions}
To examine whether MetaTSI can encode continuous functions and enable coordinate interpolation, we evaluate its continuity. 
We start by supposing that the Lipschitz constant of sine activation is $\epsilon$, all bias terms are absorbed in weight matrices, and the $\ell_1$ norm of weight matrices, the core tensor $\mathcal{C}$, as well as all latent vectors $\phi$ are bounded by $\xi$, $\lambda$, and $\eta$ respectively. Then $\forall \Phi(\tau)$ we have:
\begin{equation}
    \begin{aligned}
        |\Phi(\tau)|&\leq |\mathbf{W}^{(L+1)}||(\mathbf{W}_s^{L}\phi)\odot\sin(\mathbf{W}^{L}\cdots \\
        &\cdots(\mathbf{W}_s^{1}\phi)\odot\sin(\mathbf{W}^{1}))||\tau| \leq \epsilon^L\xi^{2L+1}\eta^L|\tau|.
    \end{aligned}
\end{equation}

Then for the factorized MetaTSI model $\Phi_{\Theta}(\bm{\vec{\tau}}) = \mathcal{C}\times_1\Phi_{\theta_1}({\vec{\tau}}^1)\times_2\Phi_{\theta_2}({\vec{\tau}}^2)$, we have: 
\begin{equation}\label{eq:lipschitz}
    \begin{aligned}
        |\Phi(\vec{\tau}^1,\vec{\tau}^2)-\Phi(\vec{\tau}^{1'},\vec{\tau}^2)|&\leq\lambda\epsilon^{2L}\xi^{4L+2}\eta^{2L}\nu|\vec{\tau}^1-\vec{\tau}^{1'}|,\\
        |\Phi(\vec{\tau}^1,\vec{\tau}^2)-\Phi(\vec{\tau}^1,\vec{\tau}^{2'})|&\leq\lambda\epsilon^{2L}\xi^{4L+2}\eta^{2L}\nu|\vec{\tau}^2-\vec{\tau}^{2'}|,\\
    \end{aligned}
\end{equation}
where $\nu=\max\{\vec{\tau}^1,\vec{\tau}^2\}$ and the above inequality holds based on the fact that $|a\odot b|\leq | a| |b |$.
The above discussion indicates that $\Phi$ is Lipschitz continuous in the input coordinate system, serving as a smooth function approximator.

\subsubsection{MetaTSI as Fourier Series}
To answer the research question of why MetaTSI is effective for time series imputation, we explore its property through Fourier analysis. We show that hidden representations in the modulated mapping network are equivalent to a composition of Fourier series, which naturally characterizes the representation of the underlying signal.

First, we indicate that the CRF in Eq. \ref{eq:crf} is essentially a sine term. Considering the simplest case that $N_f=1$, i.e.,
\begin{equation}
\begin{aligned}
    \gamma(\tau)=[&\sin(2\pi\mathbf{B}\tau),\cos(2\pi\mathbf{B}\tau)]^{\mathsf{T}}.
\end{aligned}
\end{equation}

Then it passes through the first layer of Eq. \ref{eq:freqmlp}:
\begin{equation}
\begin{aligned}
    \vh^{(1)} &= \mathbf{W}^{(0)}\gamma_{\sigma}(\tau)+\vb^{(0)}, \\
    &=\mathbf{W}^{(0)}\left[\sin(2\pi\mathbf{B}\tau),\cos(2\pi\mathbf{B}\tau)\right]^{\mathsf{T}}+\vb^{(0)}, \\
    &=\mathbf{W}^{(0)}\sin(2\pi\mathbf{B}'\tau+\rho)+\vb^{(0)},
\end{aligned}
\end{equation}
where $\mathbf{B}'=[\mathbf{B},\mathbf{B}]^{\mathsf{T}}$ and $\rho=[\pi/2,\dots,\pi/2,0,\dots,0]$. 
This shows that it is equivalent to a single-layer network with sine activation. The case $N_f\neq 1$ can be verified similarly.

Second, the output of subsequent layers can be approximated as a combination of sinusoidal bases: 
\begin{equation}
    \vh^{(\ell+1)} \approx\sum_{k=1}^K\Bar{\alpha}_k\sin(2\pi\Bar{\omega}_k \vh^{(\ell)}+\Bar{\rho}_k) + \Bar{\beta}_k,
\end{equation}
where $K$ is the total order of coefficients $\Bar{\alpha}_k$, frequencies $2\pi\Bar{\omega}_k$, phase shifts $\Bar{\rho}_k$, and bias $\Bar{\beta}_k$.
This is evidenced by:
\begin{equation}\label{eq:fourier_approx}
\begin{aligned}
    \vh^{(\ell+1)} &= \delta^{(\ell),i} \odot \sin(\mathbf{W}^{(\ell)}\vh^{(\ell)}+\vb^{(\ell)}),\\
    &= \delta^{(\ell),i} \odot \sin(\mathbf{W}^{(\ell)}(\delta^{(\ell-1),i} \odot \sin(\dots))+\vb^{(\ell)}), \\
    &=(\widetilde{\mathbf{W}}^{(\ell)}\circ\sin\circ\cdots\circ\widetilde{\mathbf{W}}^{(0)})(\tau),
\end{aligned}
\end{equation}
where the bias term and element-wise product are absorbed in a new weight matrix $\widetilde{\mathbf{W}}^{(\ell)}$.
As sinusoidal activation can be effectively approximated using polynomials with a Taylor expansion \cite{yuce2022structured}, the composition of sinusoidal signals is still a set of wave signals. This makes the hidden state resemble a Fourier representation of the underlying signal.

Last, by observing Eq. \ref{eq:fourier_approx}, we find that the modulation vector $\delta^{(\ell),i}$ affects the amplitude, phase shift, and frequency in each hidden layer of the mapping network. This property guarantees the expressivity of the latent modulation, making MetaTSI generalizable to large-scale complex datasets.

\section{Experiments}\label{sec:experiments}
This section designs a battery of experiments to evaluate whether MetaTSI can provide high-performance, generalizable, and computationally efficient imputation for urban time series. We (1) compare it with state-of-the-art imputation models on single-city benchmarks; (2) demonstrate generalization across cites and sensors; (3) assess its few-shot performance in unseen samples; and (4) detail additional properties of MetaTSI, such as computational efficiency, architectural rationality, and interpretable latent manifolds. 

\begin{table}[!htb]
\renewcommand{\arraystretch}{0.9} 
\setlength{\abovecaptionskip}{0.cm}
\setlength{\belowcaptionskip}{-0.0cm}
\caption{Statistics of cross-city datasets.}
\label{tab:data_info}
\centering
\setlength{\tabcolsep}{6pt}
\resizebox{0.75\columnwidth}{!}{
\begin{tabular}{l|c c c c}
\toprule
 City & \# of Sensors & Seq. Len. & Frequency \\
\toprule
\texttt{London} & 797 & 1440 & 5 min  \\
\texttt{Orange} & 344 & 1344 & 15 min  \\
\texttt{Utrecht}  & 407 & 1152 & 5 min  \\
\texttt{Los Angeles}  & 702 & 1344 & 3 min \\
\texttt{San Diego}  & 702 & 1344 & 15 min  \\
\texttt{Riverside}  & 474 & 1344 & 15 min  \\
\texttt{San Francisco}  & 259 & 1344 & 15 min  \\
\texttt{Melbourne}  & 926 & 672 & 15 min  \\
\texttt{Bern}  & 462 & 2015 & 5 min  \\
\texttt{Kassel}  & 292 & 1171 & 5 min  \\
\texttt{Darmstadt}  & 112 & 1936 & 3 min  \\
\texttt{Toronto}  & 184 & 672 & 15 min  \\
\texttt{Speyer}  &  63& 2016 & 3 min  \\
\texttt{Manchester}  & 148 & 1911 & 5 min  \\
\texttt{Luzern}  & 101 & 3360 & 3 min  \\
\texttt{Taipei}  & 379 & 3280 & 3 min  \\
\texttt{Contra Costa}  & 448 & 1344 & 15 min  \\
\texttt{Hamburg}  &254  & 2016 &  3 min  \\
\texttt{Munich}  & 437 & 288 & 5 min  \\
\texttt{Zurich}  &  803& 2016 & 3 min  \\
\bottomrule
\end{tabular}}
\end{table}

\subsection{Datasets and Experimental Settings}
\noindent\textbf{Datasets}. To scale the analysis and evaluation to cover multiple cities across the world, we collect and construct a large-scale urban traffic benchmark. This benchmark includes 20 cites worldwide, with traffic flow data processed from UTD19 data \footnote{\url{https://utd19.ethz.ch/}} and PeMS data \footnote{\url{https://pems.dot.ca.gov/}}. UTD19 consists mainly of measurements from loop detectors from 2017-2019, which record vehicle flow and occupancy in a relatively small aggregation interval, typically 3-5min. The cities included in UTD19 are located mainly in European countries. PeMS data is collected from individual detectors spanning the freeway system across all major metropolitan areas of California, which is processed into a regular interval of 5 minutes.
The whole dataset includes more than 8,000 heterogeneous time series with different lengths and frequencies. Brief summary of the data is given in Table. \ref{tab:data_info}.
To compare with existing models, we ensure that time series from the same city have the same length. For a straightforward evaluation, we report the MAE and MSE metrics on Z-normalized values as different series have different scales and dimensions.

\begin{table*}[!htbp]
\renewcommand{\arraystretch}{0.9} 
\setlength{\abovecaptionskip}{0.cm}
\setlength{\belowcaptionskip}{-0.0cm}
\caption{{Result (normalized metrics) in \texttt{London}, \texttt{Utrecht}, \texttt{Manchester}, \texttt{San Francisco}, \texttt{Melbourne} and \texttt{Toronto}.}}
\begin{center}
\setlength{\tabcolsep}{10pt}
\resizebox{0.95\textwidth}{!}{
\begin{tabular}{l|l|cc|cc|cc|cc|cc|cc}
\toprule
 & & \multicolumn{2}{c|}{\texttt{London}} & \multicolumn{2}{c|}{\texttt{Utrecht}} & \multicolumn{2}{c|}{\texttt{Manchester}} & \multicolumn{2}{c|}{\texttt{San Francisco}} & \multicolumn{2}{c|}{\texttt{Melbourne}} & \multicolumn{2}{c}{\texttt{Toronto}} \\
\cmidrule{2-14}
&\multicolumn{1}{l|}{Models} & \multicolumn{1}{c}{{MAE}} & \multicolumn{1}{c|}{{MSE}} & \multicolumn{1}{c}{{MAE}} & \multicolumn{1}{c|}{{MSE}}& \multicolumn{1}{c}{{MAE}} & \multicolumn{1}{c|}{{MSE}}& \multicolumn{1}{c}{{MAE}} & \multicolumn{1}{c|}{{MSE}}& \multicolumn{1}{c}{{MAE}} & \multicolumn{1}{c|}{{MSE}}& \multicolumn{1}{c}{{MAE}} & \multicolumn{1}{c}{{MSE}}\\
\hline
\multirow{13}{*}{\rotatebox{90}{$20\%$ observation rate}} &{Average} & 0.485  & 0.438  & 0.685 & 0.874 & 0.533 & 0.531 &  0.816 & 0.998 & 0.649 & 0.766 & 0.562 & 0.582 \\
&{TRMF (NeurIPS'16)} & 0.245  & 0.129  & 0.324 & 0.236 & \underline{0.139}  & \underline{0.056} & 0.157 & 0.099 & 0.245 & 0.196 &0.199  & 0.095 \\
&{TIDER (ICLR'23)} & 0.282  &  0.179 & 0.363 & 0.275 &  0.206 & 0.105 &  0.273 & 0.150 & 0.245 & 0.124 & 0.322  & 0.212 \\
&{mTAN (ICLR'21)} & 0.345  & 0.188  & 0.491 & 0.552 & 0.387 & 0.303 & 0.567 & 0.490 & 0.498 & 0.527 & 0.679 & 0.888 \\
&{SIREN (NeurIPS'20)} & 0.674  & 0.847  & 0.529 & 0.629 & 0.668 & 0.773 & 0.341 &  0.235 & 0.515 &  0.531 &  0.749 & 0.984 \\
&{FourierNet (NeurIPS'20)} & 0.627  & 0.701  & 0.334 & 0.248 & 0.155 & 0.061 &0.162  & 0.062 & 0.530 & 0.548 & 0.358 & 0.316 \\
&{DeepTime (ICML'23)} & 0.491  & 0.526  & 0.474 & 0.452 & 0.451 & 0.425 & 0.526 & 0.527 & 0.535 & 0.611 & 0.464 & 0.508 \\
&{TimeFlow (TMLR'24)} & 0.287  & 0.181  & 0.347 &0.262  & 0.255 & 0.157 &  0.347 & 0.224 & 0.311 & 0.248 & 0.393 & 0.322 \\
&{LRTFR (TPAMI'23)} & 0.256  & 0.196  &0.310  & 0.218 & 0.335 & 0.233 & 0.327 & 0.217 & 0.229 & 0.111 & 0.295 & 0.193 \\
&{LCR-1D (TKDE'24)} & {{0.267}}  &{{0.134}} &{0.320}  & 0.208 &{0.161}  &  0.064  & 0.155  &  0.051 &\underline{0.228} & \underline{0.103} &\underline{0.195} & \underline{0.085} \\
&{LCR-2D (TKDE'24)}  & 0.291  & 0.151 & 0.329  &  0.211  & 0.189 &  0.077 &\underline{0.143} & \underline{0.042}  & 0.254 & 0.121 & 0.212 & 0.092 \\
\cmidrule{2-14}
&\textbf{MetaTSI-1D} (ours)& \cellcolor{gray!15}\underline{0.234} & \cellcolor{gray!15}\underline{0.121} & \cellcolor{gray!15}\underline{0.266} & \cellcolor{gray!15}\underline{0.154} & \cellcolor{gray!15}{0.159} & \cellcolor{gray!15}{0.061} & \cellcolor{gray!15}{0.181} & \cellcolor{gray!15}{0.077} & \cellcolor{gray!15}{0.244} & \cellcolor{gray!15}{0.122} & \cellcolor{gray!15}{0.202} & \cellcolor{gray!15}{0.098}\\
&\textbf{MetaTSI-2D} (ours)& \cellcolor{gray!30}$\textbf{0.212}$ & \cellcolor{gray!30}{$\textbf{0.091}$} & \cellcolor{gray!30}{$\textbf{0.257}$} &  \cellcolor{gray!30}{$\textbf{0.148}$} &  \cellcolor{gray!30}{$\textbf{0.115}$} &  \cellcolor{gray!30}{$\textbf{0.045}$} &  
\cellcolor{gray!30}{$\textbf{0.131}$} & \cellcolor{gray!30}{$\textbf{0.040}$}& \cellcolor{gray!30}{$\textbf{0.183}$} &  \cellcolor{gray!30}{$\textbf{0.071}$} &  \cellcolor{gray!30}{$\textbf{0.159}$} &  \cellcolor{gray!30}{$\textbf{0.065}$} \\
\hline
\multirow{13}{*}{\rotatebox{90}{$10\%$ observation rate}} &{Average} & 0.486  & 0.439  & 0.684 & 0.879 &0.533  & 0.534 & 0.820 & 1.005 & 0.651 & 0.772 & 0.563 & 0.593 \\
&{TRMF (NeurIPS'16)} & 0.279  & 0.161  & 0.352 & 0.259 & 0.188 & 0.120 & \underline{0.189} & 0.096 & \underline{0.278} & 0.160 & 0.282 & 0.174 \\
&{TIDER (ICLR'23)} &  0.384  &  0.299 &0.467  & 0.420 &  0.324 & 0.224 & 0.430 & 0.343 & 0.423 & 0.335 & 0.487 & 0.467 \\
&{mTAN (ICLR'21)} & 0.352  & 0.194  & 0.390 & 0.292 & 0.411 & 0.389 & 0.593 & 0.550 & 0.564 & 0.589 & 0.701 & 0.958 \\
&{SIREN (NeurIPS'20)} & 0.692  & 0.900  & 0.534 & 0.640 &0.701  & 0.892 & 0.385 & 0.310 & 0.527 & 0.609 & 0.723 & 1.063 \\
&{FourierNet (NeurIPS'20)} & 0.648  &  0.784 & 0.382 & 0.342 & 0.261 & 0.171 &0.237  & 0.135 & 0.535 & 0.556 & 0.456 & 0.544 \\
&{DeepTime (ICML'23)} & 0.567  & 0.662  & 0.610  & 0.720 & 0.635 & 0.763 & 0.740 &  0.930 & 0.628 &0.790  & 0.542 & 0.652 \\
&{TimeFlow (TMLR'24)} & 0.316  &  0.208 & 0.362 & 0.273 & 0.288 & 0.187 & 0.430 &  0.342 & 0.340 & 0.262 & 0.336 & 0.262 \\
&{LRTFR (TPAMI'23)} & 0.270  & 0.197  & 0.323 & 0.262 & 0.343 & 0.251 & 0.354 & 0.246 & 0.237 &0.122  & 0.310 & 0.213 \\
&{LCR-1D (TKDE'24)} & {{0.293}}  &{{0.160}}   & {0.342}  & 0.226 &{0.199}  &  0.088  & 0.204  &  0.083 &{0.282} & \underline{0.151} &0.265 & 0.142 \\
&{LCR-2D (TKDE'24)}  & 0.303  &  0.165   & 0.339  &  0.217  & 0.212 &  0.092 &{0.191} & \underline{0.072}  & 0.288 & 0.152 & \underline{0.260} & \underline{0.131} \\
\cmidrule{2-14}
&\textbf{MetaTSI-1D} (ours)& \cellcolor{gray!15}\underline{0.238} & \cellcolor{gray!15}\underline{0.125} & \cellcolor{gray!15}\underline{0.281} & \cellcolor{gray!15}\underline{0.169} & \cellcolor{gray!15}\underline{0.146} & \cellcolor{gray!15}\underline{0.066} & \cellcolor{gray!15}{0.239} & \cellcolor{gray!15}{0.125} & \cellcolor{gray!15}{0.291} & \cellcolor{gray!15}{0.186} & \cellcolor{gray!15}{0.275} & \cellcolor{gray!15}{0.191}\\
&\textbf{MetaTSI-2D} (ours)& \cellcolor{gray!30}$\textbf{0.220}$ & \cellcolor{gray!30}{$\textbf{0.102}$} & \cellcolor{gray!30}{$\textbf{0.272}$} &  \cellcolor{gray!30}{$\textbf{0.192}$} &  \cellcolor{gray!30}{$\textbf{0.123}$} &  \cellcolor{gray!30}{$\textbf{0.065}$} &  
\cellcolor{gray!30}{$\textbf{0.151}$} & \cellcolor{gray!30}{$\textbf{0.058}$}& \cellcolor{gray!30}{$\textbf{0.198}$} &  \cellcolor{gray!30}{$\textbf{0.088}$} &  \cellcolor{gray!30}{$\textbf{0.185}$} &  \cellcolor{gray!30}{$\textbf{0.088}$} \\
\hline
\multirow{13}{*}{\rotatebox{90}{$5\%$ observation rate}} &{Average} & 0.488  & 0.443  & 0.688 & 0.888 & 0.531 & 0.537 & 0.824 & 1.017 & 0.657 &0.786  & 0.564 & 0.603 \\
&{TRMF (NeurIPS'16)} &  0.320 & 0.219  & 0.380 & 0.298 & 0.275 & 0.182 & 0.296 & 0.188 & 0.387 & 0.272 & 0.333  &  0.225\\
&{TIDER (ICLR'23)} & 0.454  & 0.389  & 0.562 & 0.608 & 0.533 & 0.579 & 0.674 & 0.785 & 0.629 &0.713  & 0.719 &  1.032 \\
&{mTAN (ICLR'21)} & 0.441  & 0.395  & 0.580 & 0.662 & 0.683 & 0.810 & 0.789 & 1.008 & 0.580 & 0.626 & 0.754 & 1.113 \\
&{SIREN (NeurIPS'20)} & 0.721  & 1.220  & 0.558 & 0.725 & 0.743 & 1.023 & 0.407 & 0.357 & 0.591 & 0.757 & 0.749 & 1.140 \\
&{FourierNet (NeurIPS'20)} & 0.690  & 0.937 & 0.462 & 0.530 & 0.267 & 0.210 & 0.358 & 0.286 & 0.601 & 0.783 &0.611  & 0.843 \\
&{DeepTime (ICML'23)} &  0.654 & 0.831  & 0.752 & 1.038 & 0.739 & 0.995 & 0.861 &  1.241 &0.713  & 0.984 & 0.782 & 1.234 \\
&{TimeFlow (TMLR'24)} & 0.338  & 0.238  & 0.586 &0.773  & 0.321 & 0.224 & 0.611 & 0.655 & 0.373 & 0.307 & 0.407 & 0.362 \\
&{LRTFR (TPAMI'23)} & 0.337  &  0.293 &0.376  & 0.361 &0.358  & 0.281 & 0.431 &0.359  &0.279  & 0.174 & 0.345 & 0.267 \\
&{LCR-1D (TKDE'24)} & {{0.337}}  &{{0.210}}   & {0.380}  & 0.268 &{0.251}  &  0.127  & 0.266  &  0.136 & {0.354} & 0.230 &0.364 & 0.249 \\
&{LCR-2D (TKDE'24)}  &0.335  &  0.204   & 0.371  & 0.252  & 0.252 & 0.123 &\underline{0.242} & \underline{0.113}  & \underline{0.342} & \underline{0.209} & \underline{0.324} & \underline{0.193} \\
\cmidrule{2-14}
&\textbf{MetaTSI-1D} (ours) & \cellcolor{gray!15}\underline{0.262} & \cellcolor{gray!15}\underline{0.148} & \cellcolor{gray!15}\underline{0.337} & \cellcolor{gray!15}\underline{0.245} & \cellcolor{gray!15}\underline{0.177} & \cellcolor{gray!15}\underline{0.098} & \cellcolor{gray!15}{0.353} & \cellcolor{gray!15}{0.266} & \cellcolor{gray!15}{0.395} & \cellcolor{gray!15}{0.372} & \cellcolor{gray!15}{0.397} & \cellcolor{gray!15}{0.359}\\
&\textbf{MetaTSI-2D} (ours) & \cellcolor{gray!30}$\textbf{0.244}$ & \cellcolor{gray!30}{$\textbf{0.144}$} & \cellcolor{gray!30}{$\textbf{0.306}$} &  \cellcolor{gray!30}{$\textbf{0.298}$} &  \cellcolor{gray!30}{$\textbf{0.139}$} &  \cellcolor{gray!30}{$\textbf{0.095}$} &  
\cellcolor{gray!30}{$\textbf{0.182}$} & \cellcolor{gray!30}{$\textbf{0.083}$}& \cellcolor{gray!30}{$\textbf{0.243}$} &  \cellcolor{gray!30}{$\textbf{0.146}$} &  \cellcolor{gray!30}{$\textbf{0.227}$} &  \cellcolor{gray!30}{$\textbf{0.136}$} \\
\bottomrule
\end{tabular}}
\end{center}
\label{tab:single_city}
\end{table*}

\noindent\textbf{Baselines}. We compare MetaTSI with both analytical models and deep implicit representation models. They are from state-of-the-art venues in related literature.
For single-city comparison, we consider: (1) SIREN (NeurIPS'20) \cite{SIREN}; (2) FourierNet (NeurIPS'20) \cite{FourierFeature}; (3) DeepTime (ICML'23) \cite{woo2023learning}; (4) TimeFlow (TMLR'24) \cite{naour2023time}; (5) LCR (TKDE'24) \cite{chen2024laplacian}; (6) TRMF (NeurIPS'16) \cite{yu2016temporal}; (7) TIDER (ICLR'23) \cite{liu2023multivariate}; (8) mTAN (ICLR'21) \cite{shukla2021multi}; (9) LRTFR (TPAMI'23) \cite{luo2023low}. For cross-city experiments, we consider generalizable models: (1) TimeFlow; (2) Functa (ICML'22) \cite{functa}; (3) DeepTime; (4) HyperNet-SIREN; (5) SIREN+.
Note that since the data has large spatial dimensions and irregular sampling frequencies, many deep imputation models (e.g., BRITS, SAITS, and ImputeFormer) are not well suited for this challenging task and we do not include them as baselines. In addition, both LCR and MetaTSI have two variations according to model implementations: (1) 1D model: treating each time series as individuals and modeling independently; (2) 2D model: treating the multivariate time series as a whole matrix and adding an additional spatial dimension.

\noindent\textbf{Hyperparameters}. There are several key hyperparameters in MetaTSI. For CRF, the number of Fourier features
$N_f$ is set to 2048, the initial scales are set to $[0.01,0.1,1,5,10,20,50,100]$, and the layerwise frequency scales are set to $[10,5,1,0.1,0.01]$. For the mapping and modulation network, the layer number is set to 5 with a hidden dimension of 1024 and the delay embedding dimension is chosen from $[24,36,72]$. For optimization, the Adam optimizer is adopted with an outer learning rate of $5\times 10^{-4}$ and an inner learning rate of $1\times 10^{-2}$. The dimension of the latent code is set to 128 and the number of inner steps is $3\sim 5$.
For baseline models, as few of them were originally designed for irregular time series imputation, we determine their respective hyperparameters using cross-validation.

\subsection{Model Comparison within a Single City}\label{results:traffic}
In the first experiment, we train each model in each city with varying observation rates (i.e., $5\%, 10\%, 20\%$). According to their mechanisms, models either are fitted to the observations and predict missing values according to the coordinates, or are optimized to reconstruct the distributions or patterns of the observations. Results in Table. \ref{tab:single_city} indicate that the proposed MetaTSI-2D model consistently achieves better performances than baselines in all cities, and MetaTSI-1D also achieves performance comparable to the state-of-the-art. When there are fewer observations, the superiority of MetaTSI is more significant.
In particular, MetaTSI-2D performs better than MetaTSI-1D in fitting observations within a single city. This is achieved by the strong spatial inductive bias imposed on the model, which leads to a larger model capacity to fit data distributions. However, in the next section we show that this sacrifices model generalizability.

\begin{table*}[!htbp]
\renewcommand{\arraystretch}{0.8} 
\setlength{\abovecaptionskip}{0.cm}
\setlength{\belowcaptionskip}{-0.0cm}
\caption{{Cross-city learning results of different models (normalized MAE and MSE).}}
\begin{center}
\setlength{\tabcolsep}{10pt}
\resizebox{0.9\textwidth}{!}{
\begin{tabular}{l|>{\columncolor{gray!30}}c|>{\columncolor{gray!30}}c|>{\columncolor{gray!15}}c|>{\columncolor{gray!15}}c|c|c|c|c|c|c}
\toprule
\multicolumn{1}{l|}{Models} & \multicolumn{2}{c|}{\textbf{MetaTSI-1D}} & \multicolumn{2}{c|}{\textbf{MetaTSI-2D}} & \multicolumn{2}{c|}{TimeFlow} & \multicolumn{2}{c|}{SIREN+}  & \multicolumn{2}{c}{Functa} \\
\hline
\multicolumn{1}{l|}{Cities} & \multicolumn{1}{c|}{\scalebox{0.9}{MAE}} & \multicolumn{1}{c|}{MSE} & \multicolumn{1}{c|}{\scalebox{0.9}{MAE}} & \multicolumn{1}{c|}{MSE}& \multicolumn{1}{c|}{\scalebox{0.9}{MAE}} & \multicolumn{1}{c|}{MSE}& \multicolumn{1}{c|}{\scalebox{0.9}{MAE}} & \multicolumn{1}{c|}{MSE}& \multicolumn{1}{c|}{\scalebox{0.9}{MAE}} & \multicolumn{1}{c}{MSE} \\
\hline
\texttt{London}  & \textbf{0.253} & \textbf{0.132} & 0.302 & 0.171 & 0.516  & 0.499 & 0.488 & 0.443 &0.514 &0.500 \\
\texttt{Orange}  & \textbf{0.238}  & \textbf{0.108} & 0.240 & 0.103 & 0.514  & 0.424 & 0.864 & 1.125&0.373 & 0.236\\
\texttt{Utrecht}  & \textbf{0.278} & \textbf{0.164} & 0.332 & 0.244 & 0.730  & 1.026 &0.667 &0.836 &0.734 & 1.057\\
\texttt{Los Angeles}  & \textbf{0.184} & \textbf{0.074} & 0.233 & 0.107 &  0.518 & 0.439 & 0.804&0.977 & 0.410 &0.282 \\
\texttt{San Diego} & \textbf{0.174} & \textbf{0.066} & 0.265 & 0.135 & 0.508  & 0.423 & 0.795 & 0.952 &0.421  & 0.295\\
\texttt{Riverside} & \textbf{0.182} & \textbf{0.073} & 0.209 & 0.091 & 0.447  & 0.332 &0.714 &0.774 & 0.351 & 0.209 \\
\texttt{San Francisco}  & \textbf{0.242} & \textbf{0.114} & 0.294 & 0.164 & 0.548  & 0.480 & 0.811 &0.991 & 0.456 & 0.349\\
\texttt{Melbourne}  & \textbf{0.329} & \textbf{0.243} & 0.360 & 0.247 & 0.484  & 0.449 &0.640 &0.748 &0.411 & 0.337\\
\texttt{Bern} & \textbf{0.291}  &  \textbf{0.178} & 0.327 & 0.220 & 0.781  & 1.096 &0.801 &1.079 & 0.730 & 1.014\\
\texttt{Kassel}   & \textbf{0.329} & \textbf{0.268} & 0.403 & 0.363 & 0.728 & 0.979  & 0.705 &0.881 &0.720 & 0.976 \\
\texttt{Darmstadt} & \textbf{0.389} & \textbf{0.327}  & 0.418 & 0.348 & 0.712 & 0.883 & 0.733&0.888 & 0.696&0.863 \\
\texttt{Toronto} & \textbf{0.367} & \textbf{0.298} & 0.322 & 0.211 & 0.506 & 0.474 &0.604 & 0.647& 0.479& 0.437 \\
\texttt{Speyer} & \textbf{0.455} & \textbf{0.387} &  0.443 & 0.344 & 0.873 & 1.306 &0.943 & 1.410 &0.834 & 1.229\\
\texttt{Manchester} & \textbf{0.173} & \textbf{0.067} & 0.196 & 0.094 & 0.607 & 0.698 & 0.515& 0.487& 0.666& 0.813\\
\texttt{Luzern} & \textbf{0.341} & \textbf{0.203} & 0.340 & 0.204 &  0.940  & 1.400 & 0.933 &1.288 & 0.962& 1.482\\
\midrule
Average & \textbf{0.282} & \textbf{0.181} & 0.313 &0.204 & 0.627 &0.727 & 0.734 & 0.902 & 0.584 &0.672 \\
\bottomrule
\end{tabular}}
\label{tab:mutiple_cities}
\end{center}
\end{table*}

\subsection{Generalization across Multiple Cities}\label{subsec:cross-city}
Second, we examine the generalizability of models. Our objective is to learn a unified model to reconstruct all time series simultaneously for all cities. We select 15 cities for this experiment and the data missing rate for each city is randomly sampled from $[80\%,95\%]$. As data are irregularly sampled in different cities, we need continuous models to deal with the irregularity.
Therefore, only continuous models equipped with transfer-learning-related strategies can complete this task. Results of different methods are given in Table. \ref{tab:mutiple_cities}. It is observed that in this case MetaTSI-2D is surpassed by MetaTSI-1D which has better generalizability. Different cities have heterogeneous spatial patterns such as different network topologies and mobility demands. Learning a city as a whole increases the challenges of model-based transfer.
Other meta-learning-based INR models cannot achieve competitive performances, which demonstrate the effectiveness of our multi-scale architecture and modulation strategy particularly designed for urban time series.


\subsection{Generalization on Unseen Instances}
Next, we evaluate the out-of-distribution generalization performance of models. This scenario happens often in practice when the pretrained model is applied to new locations or cities with limited computational resources to retrain the entire model. The model trained on the source data is only fine-tuned by the inner loop of the meta-learning scheme with few-shot observations. This is accompanied by a small number of gradient adaptation steps, e.g., 5 in our experiments.
We compare MetaTSI with state-of-the-art implicit time series imputation model TimeFlow.

\noindent\textbf{Generalization on unseen series}.
We split the data used in Section \ref{subsec:cross-city} into two separate sets, each containing half of the series.
We pretrain the model on half of the data and fine-tune it with the meta-learning adaptation (lines 7-10 in Algorithm \ref{algo:meta-learning}) for the other half set. Table \ref{tab:unseen_cities_series} shows the result. As can be seen, MetaTSI-1D outperforms the other two models by a large margin, indicating that it learns representations to generalize rather than memorize.

\noindent\textbf{Generalization on unseen cities}. 
We then apply models to completely unseen cities that are never shown in the training set. The remaining five cities are adopted for evaluation. As new cities can have patterns different from source cities, this task is very challenging. Similarly, MetaTSI-1D still achieves desirable performance compared to its counterparts, implying great potentials for cold-start problems.

\begin{table}[!htbp]
\renewcommand{\arraystretch}{0.85} 
\setlength{\abovecaptionskip}{0.cm}
\setlength{\belowcaptionskip}{-0.0cm}
\caption{{Generalization on unseen series and new cities.}}
\begin{center}
\setlength{\tabcolsep}{6pt}
\resizebox{\columnwidth}{!}{
\begin{tabular}{l|l|>{\columncolor{gray!30}}c|>{\columncolor{gray!30}}c|>{\columncolor{gray!15}}c|>{\columncolor{gray!15}}c|c|c}
\toprule
&\multicolumn{1}{l|}{Models} & \multicolumn{2}{c|}{\textbf{MetaTSI-1D}} & \multicolumn{2}{c|}{\textbf{MetaTSI-2D}} & \multicolumn{2}{c}{TimeFlow} \\
\hline
&\multicolumn{1}{l|}{Cities} & \multicolumn{1}{c|}{\scalebox{0.9}{MAE}} & \multicolumn{1}{c|}{MSE} & \multicolumn{1}{c|}{\scalebox{0.9}{MAE}} & \multicolumn{1}{c|}{MSE}& \multicolumn{1}{c|}{\scalebox{0.9}{MAE}} & \multicolumn{1}{c}{MSE}  \\
\hline
\multirow{15}{*}{\rotatebox{90}{Unseen Series}}&\texttt{London}  & \textbf{0.242} & \textbf{0.125} & 0.868 & 1.285 & 0.493  & 0.468 \\
&\texttt{Orange}  & \textbf{0.239}  & \textbf{0.108} & 0.448 & 0.360 & 0.421  & 0.294 \\
&\texttt{Utrecht}  & \textbf{0.289} & \textbf{0.180} & 0.686 & 0.870 & 0.776  & 1.154 \\
&\texttt{Los Angeles}  & \textbf{0.186} & \textbf{0.076} & 0.555 & 0.523 &  0.446  & 0.335 \\
&\texttt{San Diego} & \textbf{0.176} & \textbf{0.067} & 0.552 & 0.522 & 0.468  & 0.365 \\
&\texttt{Riverside} & \textbf{0.181} & \textbf{0.073} & 0.579 & 0.589 & 0.375  & 0.245    \\
&\texttt{San Francisco}  &  \textbf{0.256} & \textbf{0.132} & 0.522 & 0.461 & 0.498  & 0.411 \\
&\texttt{Melbourne}  &  \textbf{0.341} & \textbf{0.257} & 0.754 & 0.955 & 0.439  & 0.374  \\
&\texttt{Bern} & \textbf{0.297}  &  \textbf{0.179} & 0.581 & 0.635 & 0.758 & 1.056  \\
&\texttt{Kassel}   & \textbf{0.336} & \textbf{0.254} & 0.668 & 0.828 & 0.775 & 1.074  \\
&\texttt{Darmstadt} & \textbf{0.410} & \textbf{0.348}  &0.755 & 0.981 & 0.746 & 0.934  \\
&\texttt{Toronto} & \textbf{0.330}  & \textbf{0.238} & 0.872 & 1.245 & 0.442 & 0.378  \\
&\texttt{Speyer} & \textbf{0.454} &  \textbf{0.387} &  0.549 & 0.522 &  0.862 & 1.255  \\
&\texttt{Manchester} & \textbf{0.178} & \textbf{0.082} & 0.886 & 1.383 & 0.657  & 0.822 \\
&\texttt{Luzern} & \textbf{0.338} & \textbf{0.200} & 0.571 & 0.579 & 0.896  & 1.281 \\
\midrule
& Average & \textbf{0.284} & \textbf{0.180} & 0.656 & 0.783 & 0.603  & 0.696 \\
\midrule
\multirow{3}{*}{\rotatebox{90}{New Cities}}
&\texttt{Taipei} & \textbf{0.435} & \textbf{0.375} & 0.910 & 1.437 & 0.631 & 0.746 \\
&\texttt{Contra Costa} & \textbf{0.223} & \textbf{0.098} & 0.592 & 0.610 & 0.561 & 0.513 \\
&\texttt{Hamburg} &\textbf{0.433} &\textbf{0.363}  & 0.776 & 1.035 & 0.740 &0.997  \\
&\texttt{Munich} &\textbf{0.266} & \textbf{0.189} & 0.678 & 0.815 & 0.583 & 0.701 \\
&\texttt{Zurich} &\textbf{0.416} & \textbf{0.332} & 0.752 & 0.997 &0.721  & 0.947 \\
\midrule
& Average &\textbf{0.354} &\textbf{0.271} & 0.742 & 0.979 &0.647  & 0.781 \\
\bottomrule
\end{tabular}}
\label{tab:unseen_cities_series}
\end{center}
\end{table}

\subsection{Model Efficiency and Robustness}
This section details additional properties of MetaTSI, including both its computational efficiency and robustness.

\subsubsection{Inference Efficiency}
\begin{figure}[!htbp]
  \centering
  \captionsetup{skip=1pt}
  \includegraphics[width=1\columnwidth]{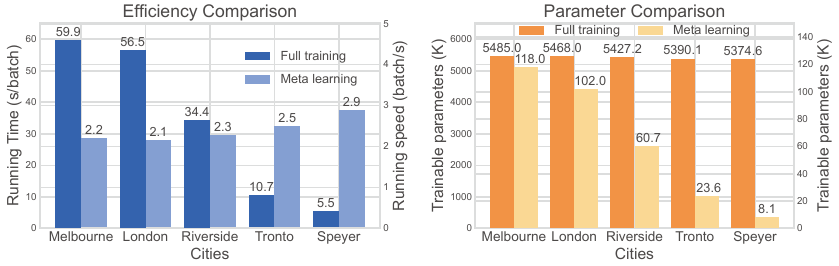}
  \caption{Efficiency and parameter study. Note that different axes have different scales and labels for better visualization.}
  \label{fig:efficiency}
\end{figure}

A salient property of MAML is its efficiency in adapting to different instances. We compare it with the full-training strategy when applied to all series within a single city. Figure \ref{fig:efficiency} compares the training speed (time) and the parameter count for the two strategies. It is evident that meta-learning significantly reduces computational expenses and shows better scalability and parameter efficiency.

\begin{table*}[!htbp]
\renewcommand{\arraystretch}{0.9} 
\setlength{\abovecaptionskip}{0.cm}
\setlength{\belowcaptionskip}{-0.0cm}
\caption{Model generalization on new time series (unseen) with varying missing rates (in terms of MAE). \textbf{MetaTSI only performs meta learning procedure with a small number (3-5 steps) of adaptation steps}.}
\centering
\setlength{\tabcolsep}{8pt}
\resizebox{0.9\textwidth}{!}{
\begin{tabular}{l|c|c|c|c|c|c|c|c|c|c|c|c}
\toprule
 New series & \multicolumn{3}{c|}{\texttt{San Diego}} & \multicolumn{3}{c|}{\texttt{Riverside}}   & \multicolumn{3}{c|}{\texttt{Orange}} & \multicolumn{3}{c}{\texttt{Los Angeles}}\\
\hline
\multirow{2}{*}{Models} & \multicolumn{3}{c|}{Observed rate} & \multicolumn{3}{c|}{Observed rate} & \multicolumn{3}{c|}{Observed rate} & \multicolumn{3}{c}{Observed rate} \\
\cmidrule(lr){2-13} 
& \multicolumn{1}{c|}{$10\%$} & \multicolumn{1}{c|}{\texttt{$5\%$}} & \multicolumn{1}{c|}{$1\%$} & \multicolumn{1}{c|}{$10\%$} & \multicolumn{1}{c|}{\texttt{$5\%$}} & \multicolumn{1}{c|}{$1\%$} & \multicolumn{1}{c|}{$10\%$} & \multicolumn{1}{c|}{\texttt{$5\%$}} & \multicolumn{1}{c|}{$1\%$} & \multicolumn{1}{c|}{$10\%$} & \multicolumn{1}{c|}{\texttt{$5\%$}} & \multicolumn{1}{c}{$1\%$}\\
\hline
{FourierNet$^\dag$ } &0.254  &  0.274 & \underline{0.565} & 0.264 & 0.268 & 0.501 & \underline{0.212} & 0.339 & 0.589 & 0.254 & \underline{0.271} & \underline{0.492} \\
{LRTFR$^\dag$}  & \underline{0.227} & \underline{0.235} & 0.597 & 0.204 & 0.289 & \underline{0.471} & 0.270 & \underline{0.316} & \underline{0.581} & \underline{0.227} & 0.330 & 0.513\\
{TimeFlow$^\ddag$} & 0.389 & 0.505 & 0.759 & 0.320 & 0.424 & 0.695 & 0.479 & 0.661 & 0.876 & 0.391 & 0.568 & 0.764\\
{TimeFlow$^\dag$} & 0.403 & 0.468 & 0.782 & 0.390 & 0.398 & 0.726 & 0.340 & 0.513 & 0.856 & 0.404 & 0.487 & 0.802\\
{Functa$^\ddag$} & 0.301 & 0.385 & 0.871 & 0.279 & 0.383 & 0.864 & 0.313 & 0.431 & 0.976 & 0.297 & 0.382 & 0.873 \\
{Functa$^\dag$} & 0.332 & 0.391 & 0.922 & 0.269 & 0.401 & 0.929 & 0.308 & 0.426 & 0.981 & 0.311 & 0.420 & 0.965 \\
{MetaTSI$^\dag$}  & 0.229 & 0.308 &0.630  & \underline{0.192} & \underline{0.255}  & 0.589  & 0.260 & 0.358 & 0.748 & 0.230 & 0.310 & 0.591 \\
\hline
\textbf{MetaTSI} &\cellcolor{gray!30}\textbf{0.162} & \cellcolor{gray!30}\textbf{0.226} & \cellcolor{gray!30}\textbf{0.437} & \cellcolor{gray!30}\textbf{0.164} & \cellcolor{gray!30}\textbf{0.248} & \cellcolor{gray!30}\textbf{0.449}  & \cellcolor{gray!30}\textbf{0.178} & \cellcolor{gray!30}\textbf{0.250} & \cellcolor{gray!30}\textbf{0.493} & \cellcolor{gray!30}\textbf{0.162} & \cellcolor{gray!30}\textbf{0.223} & \cellcolor{gray!30}\textbf{0.374}\\
\bottomrule
\end{tabular}}
\begin{tablenotes}
\item  \footnotesize{$^\dag$: Individual models for each city trained from scratch using the observed data.}
\item  \footnotesize{$^\ddag$: A unified model trained from scratch on the union of all data available.}
\end{tablenotes}
\label{tab:varying_missing}
\end{table*}

\begin{table*}[!htbp]
\renewcommand{\arraystretch}{0.9} 
\setlength{\abovecaptionskip}{0.cm}
\setlength{\belowcaptionskip}{-0.0cm}
\caption{Model generalization on new cities (unseen) with varying missing rates (in terms of MAE). \textbf{MetaTSI only performs meta learning procedure with a small number (3-5 steps) of adaptation steps}. }
\centering
\setlength{\tabcolsep}{8pt}
\resizebox{0.9\textwidth}{!}{
\begin{tabular}{l|c|c|c|c|c|c|c|c|c|c|c|c}
\toprule
 New cities & \multicolumn{3}{c|}{\texttt{Taipei}} & \multicolumn{3}{c|}{\texttt{Contra Costa}}   & \multicolumn{3}{c|}{\texttt{Hamburg}} & \multicolumn{3}{c}{\texttt{Munich}}\\
\hline
\multirow{2}{*}{Models} & \multicolumn{3}{c|}{Observed rate} & \multicolumn{3}{c|}{Observed rate} & \multicolumn{3}{c|}{Observed rate} & \multicolumn{3}{c}{Observed rate} \\
\cmidrule(lr){2-13} 
& \multicolumn{1}{c|}{$10\%$} & \multicolumn{1}{c|}{\texttt{$5\%$}} & \multicolumn{1}{c|}{$1\%$} & \multicolumn{1}{c|}{$10\%$} & \multicolumn{1}{c|}{\texttt{$5\%$}} & \multicolumn{1}{c|}{$1\%$} & \multicolumn{1}{c|}{$10\%$} & \multicolumn{1}{c|}{\texttt{$5\%$}} & \multicolumn{1}{c|}{$1\%$} & \multicolumn{1}{c|}{$10\%$} & \multicolumn{1}{c|}{\texttt{$5\%$}} & \multicolumn{1}{c}{$1\%$}\\
\hline
{SIREN+$^\ddag$} & 0.566 &  0.611 & 0.709 & 0.772 & 0.766 & 0.897 & \underline{0.592} & \underline{0.658} & 0.864 & 0.391 & \underline{0.444} &0.770 \\
{TimeFlow$^\ddag$} & \underline{0.484} &  \underline{0.583} & \underline{0.596} & \underline{0.469} & \underline{0.680}  & 0.908 & 0.596 & 0.688 & \textbf{0.742} & \underline{0.340} & 0.516 & 0.703\\
{Functa$^\ddag$} & 0.592 & 0.589  & 0.600 & 0.817 & 0.813 & \underline{0.834} & 0.736 & 0.738 & \underline{0.752} & 0.498 & 0.500 & \underline{0.681} \\
\hline
\textbf{MetaTSI} & \cellcolor{gray!30}\textbf{0.442} & \cellcolor{gray!30}\textbf{0.493} & \cellcolor{gray!30}\textbf{0.586} & \cellcolor{gray!30}\textbf{0.257} & \cellcolor{gray!30}\textbf{0.288} & \cellcolor{gray!30}\textbf{0.410} & \cellcolor{gray!30}\textbf{0.531} & \cellcolor{gray!30}\textbf{0.619} & \cellcolor{gray!30}0.792 & \cellcolor{gray!30}\textbf{0.283} & \cellcolor{gray!30}\textbf{0.413} & \cellcolor{gray!30}\textbf{0.631}\\
\bottomrule
\end{tabular}}
\begin{tablenotes}
\item  \footnotesize{$^\ddag$: A unified model trained from scratch on the union of all data available.}
\end{tablenotes}
\label{tab:varying_missing_new_cities}
\end{table*}

\subsubsection{Robustness with Sparser Data}
To justify the necessity of the collaborative imputation scheme developed in this paper, we evaluate the model robustness under sparse data conditions in Tables \ref{tab:varying_missing} and \ref{tab:varying_missing_new_cities}. The pretrained MetaTSI efficiently adapts to the new tasks by exploiting the inherent knowledge in weights of the base network.
The baselines are either individual models trained from scratch using limited observations or a unified model trained on all available data at the same time. Results reveal that in scenarios where the target city has very limited observations (e.g., only $1\%$) available, cross-city collaboration can significantly achieve performance gains.

\subsection{Ablation Study}
To justify the rationality of each modular design, we perform architectural ablation studies in Tab. \ref{tab:ablation} using two cities. They include the following architecture variations: (1) replacing the initialization scheme of latent code; (2) replacing the hierarchical modulation scheme with a fixed phase modulation used in \cite{naour2023time}; (3) replacing sinusoidal activation; (4) removing the meta-learning procedure; (5) removing the local loss; (6) removing the coordinate delay embedding; (7) removing the masked instance normalization; (8) removing CRF. According to the evaluation results, each modular design contributes to overall performance.

\begin{table}[!htbp]
\renewcommand{\arraystretch}{0.9} 
\setlength{\abovecaptionskip}{0.cm}
\setlength{\belowcaptionskip}{-0.0cm}
\caption{Ablations studies.}
\label{tab:ablation}
\centering
\setlength{\tabcolsep}{5pt}
\resizebox{0.9\columnwidth}{!}{
    \begin{tabular}{c|c|cc|cc}
    \toprule
    & \multicolumn{1}{c|}{\multirow{2}{*}{Variation}} & \multicolumn{2}{c|}{\texttt{London}} & \multicolumn{2}{c}{\texttt{S.F}}\\
    \cmidrule(lr){3-5} \cmidrule(lr){5-6} 
    &  & MAE & MSE & MAE & MSE   \\
    \midrule
     \textbf{Full} & \cellcolor{gray!30}MetaTSI-1D & \cellcolor{gray!30}\textbf{0.234} & \cellcolor{gray!30}\textbf{0.121} & \cellcolor{gray!30}\textbf{0.181} & \cellcolor{gray!30}\textbf{0.077} \\
     \midrule
     \multirow{3}{*}{Replace}& Zero init. $\rightarrow$ Random init. & 0.252 & 0.157  & 0.226 &0.110\\
     & Hierarchical $\rightarrow$ Fixed  & 0.318 &  0.223  & 0.814 & 0.989 \\
     & Sine $\rightarrow$ ReLU  & 0.246 &  0.135 & 0.236 & 0.126 \\
    \midrule
    \multirow{5}{*}{w/o} & Meta learning & 0.289 &  0.199 & 0.312 & 0.200\\
     & Variation loss & 0.250 & 0.144  & 0.248 & 0.133\\
     &  Delay emb. & 0.523 & 0.558  & 0.436 & 0.373 \\
     & Masked instance norm.  & 0.578 & 0.721  & 0.506 & 0.455\\
     & CRF  & 0.244 & 0.130  & 0.240 & 0.118\\
    \bottomrule
    \end{tabular}}
\end{table}

\subsection{Hyperparameter Study}
\begin{figure}[!htbp]
\centering
\begin{subfigure}[b]{0.20\textwidth}
\includegraphics[scale=0.28]{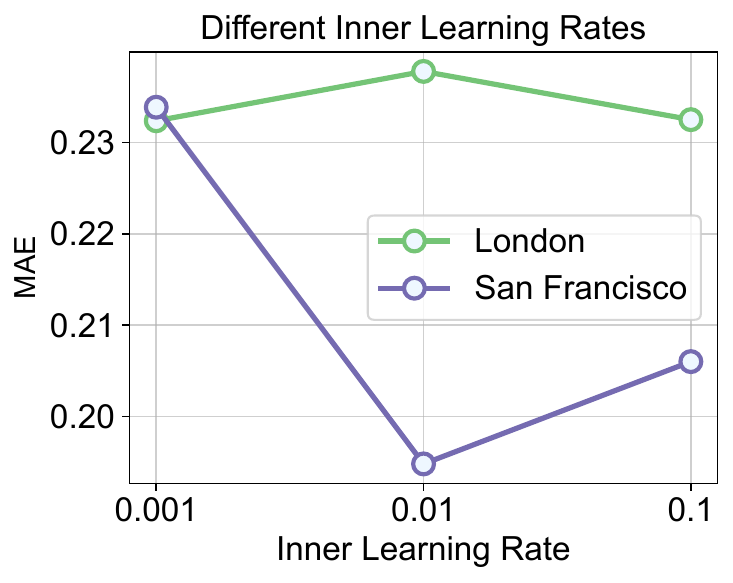}
\caption{Inner learning rates}
\end{subfigure}
\begin{subfigure}[b]{0.20\textwidth}
\centering
\includegraphics[scale=0.28]{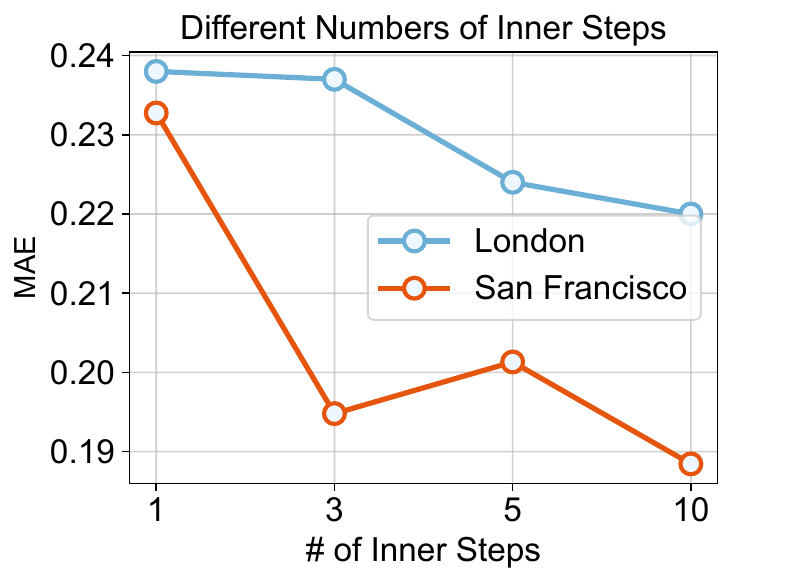}
\caption{Inner steps}
\end{subfigure}
\begin{subfigure}[b]{0.20\textwidth}
\centering
\includegraphics[scale=0.27]{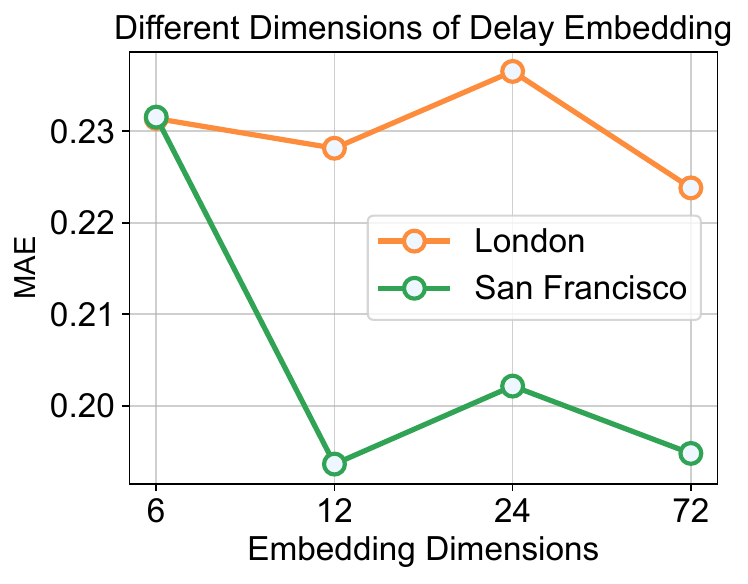}
\caption{Dim. of delay lag}
\end{subfigure}
\begin{subfigure}[b]{0.20\textwidth}
\centering
\includegraphics[scale=0.27]{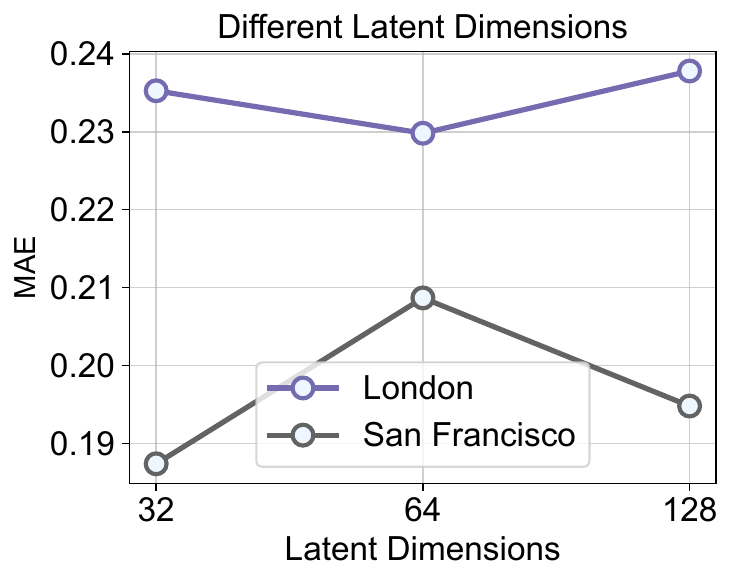}
\caption{Dim. of latent code}
\end{subfigure}
\caption{Hyperparameter studies.}
\label{fig:hyperparameter}
\end{figure}

Figure \ref{fig:hyperparameter} examines the effects of several hyperparameters, including the number of inner steps, the inner learning rate, the dimensions of the delay embedding $\delta$ and the latent coding. There are several observations: (1) a larger inner learning rate encourages learning in data with diverse instances; (2) a small number of inner steps with a proper dimension are enough to differentiate different instances; (3) a larger $\delta$ can boost performance, but increase complexity.

\subsection{Exploring the Latent Manifold}
The meta learning procedure enables a latent code learned from data to represent the series instance. In this section, we interpret these encodings using visualization tools.

\subsubsection{t-SNE Structure.}
\begin{figure}[!htbp]
  \centering
  \captionsetup{skip=1pt}
  \includegraphics[width=0.85\columnwidth]{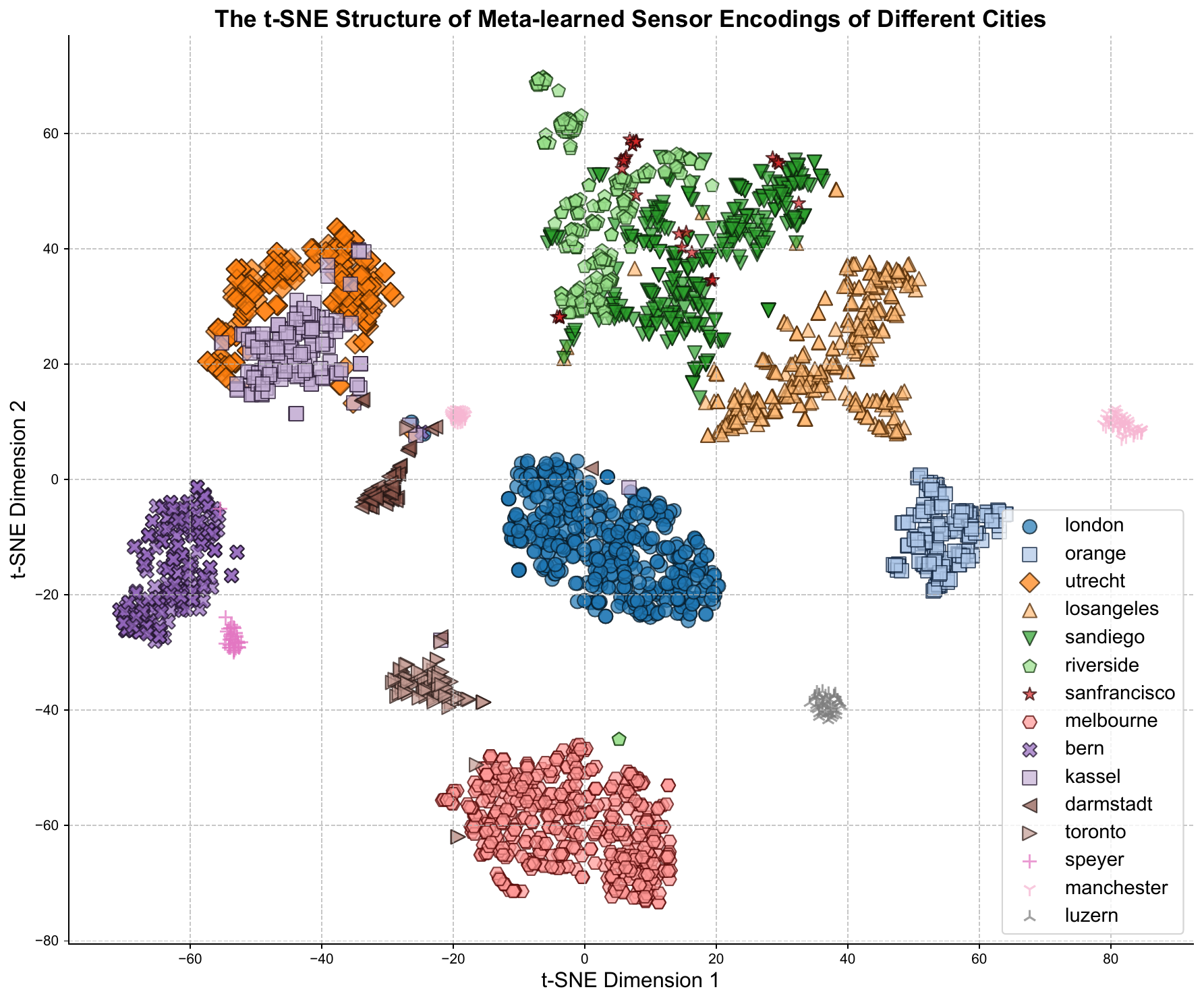}
  \caption{The t-SNE structure of meta-learned encoding.}
  \label{fig:metaenc_viz}
\end{figure}

After auto-decode the latent code for each time series in all cities, we store these codes and project them into the two-dimensional plane using the t-SNE method. Figure \ref{fig:metaenc_viz} shows the encodings of different cities. Intriguingly, MetaTSI generally separates different cities, even without any geolocation priors. Moreover, cities with similar traffic flow patterns appear in clusters that are close to each other, e.g., cities from the bay area of CA, USA.

\subsubsection{Generating Novel Time Series.}
\begin{figure}[!htbp]
  \centering
  \captionsetup{skip=1pt}
  \includegraphics[width=1\columnwidth]{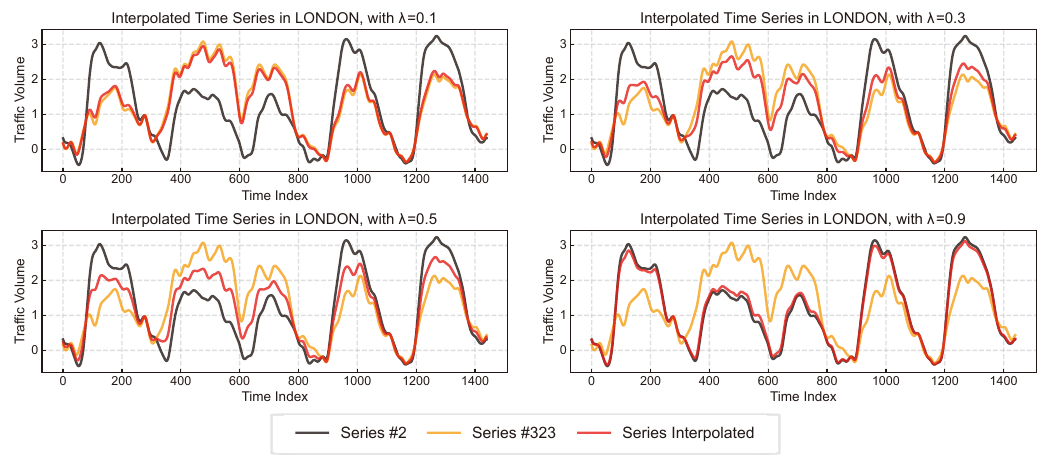}
  \caption{Decoding a time series by interpolating latent codes.}
  \label{fig:metaenc_interpolation}
\end{figure}

Recall that ones can control the output of INRs by manipulating the latent manifold.  
Since we have learned priors over the function space of all training samples, novel time series can be generated by interpolating the latent code. We examine the transitional behaviors reflected by the latent space by decoding an interpolated latent code $\Phi(\cdot;\phi_{\mu})$ with $\phi_{\mu}=\mu\phi_{1}+(1-\mu)\phi_{2}$. Figure \ref{fig:metaenc_interpolation} shows different generated time series by varying $\mu$. We observe that the interpolation path between two codes yields a ``linear'' transition in the time domain. This suggests that the latent space is smooth and well structured, which sheds light on the possibility of applying learned representations to generative modeling and extrapolation of adjacent sensors.

\subsection{Case Study: Imputation Visualization}
\begin{figure}[!htbp]
\centering
\begin{subfigure}[b]{0.98\columnwidth}
\includegraphics[scale=0.47]{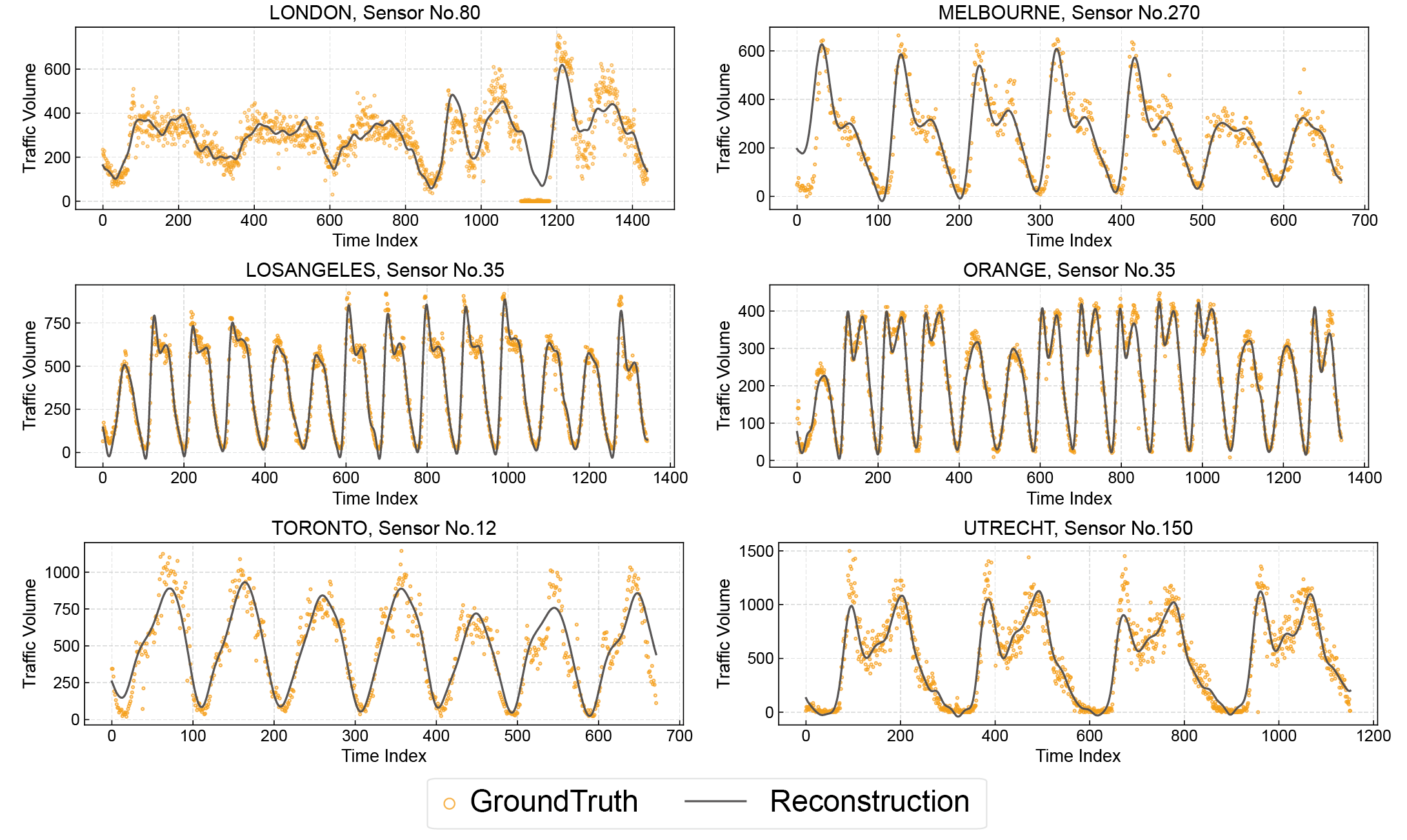}
\caption{Visualization examples of reconstructed time series.}
\end{subfigure}
\begin{subfigure}[b]{0.98\columnwidth}
\centering
\includegraphics[scale=0.47]{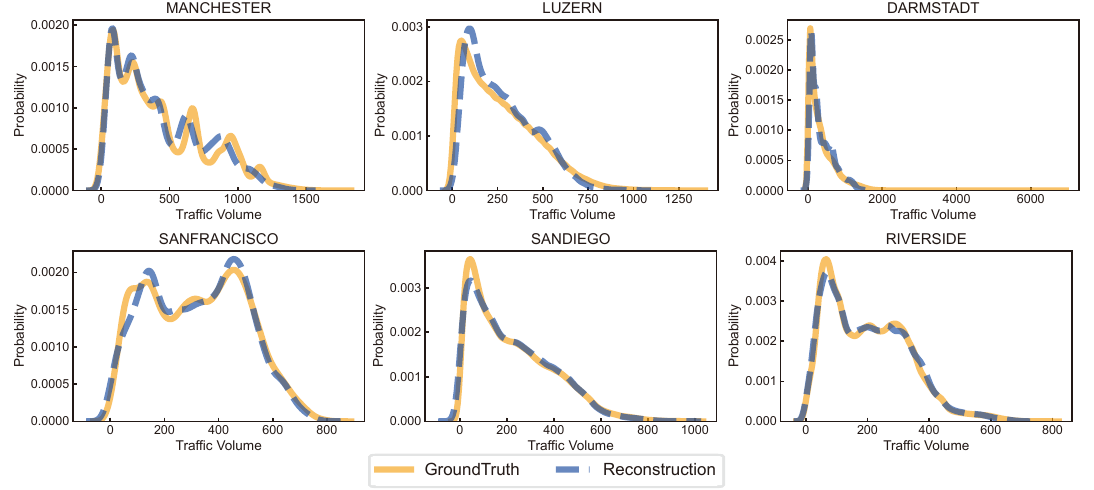}
\caption{Visualization examples of reconstructed distribution.}
\end{subfigure}
\caption{Visualization examples of imputation results.}
\label{fig:viz}
\end{figure}

Figure \ref{fig:viz} (a) displays examples of the time series imputation results in different cities and (b) shows the histogram of the true and predicted values (probability). These plots clearly show that although different cities feature distinct traffic flow observations, the shared patterns make transfer between cities possible. Our model captures these common structures and accurately reconstructs the ground truth.

\section{Conclusion and Future Direction}\label{sec: conclusion}
In this study, we introduce a novel time series imputation framework leveraging meta-learning implicit neural representations called MetaTSI.
Thanks to the generic architecture, MetaTSI originally pretrained to distinguish different instances can then generalize to heterogeneous cities to achieve collaborative imputation tasks that vary in spatiotemporal resolutions and observation patterns.
Experimental results indicate that MetaTSI not only achieves superior imputation accuracy but also excels in generalizability across diverse tasks. 
In addition, MetaTSI recovers the global structure of the signal manifold, allowing easy interpolation between nearby signals. 
Although promising results are presented, it revealed some limitations that need future efforts. First, while meta-learning facilitates efficient adaptation of new samples, pretraining it in the source data is computationally expensive and less stable than standard supervised learning. Second, this work only focuses on a single data modality. Extending MetaTSI to diverse modalities of urban data requires more sophisticated architectures.

\section*{Acknowledgments}
This research was sponsored by the National Natural Science Foundation of China (52125208), the National Natural Science Foundation of China's Fundamental Research Program for Young Students (524B2164), and a grant from the Research Grants Council of the Hong Kong Special Administrative Region, China (Project No. PolyU/15227424).





\bibliographystyle{IEEEtran}
\bibliography{references}

\vspace{-30pt}
\begin{IEEEbiography}[{\includegraphics[width=1in,height=1.25in,keepaspectratio]{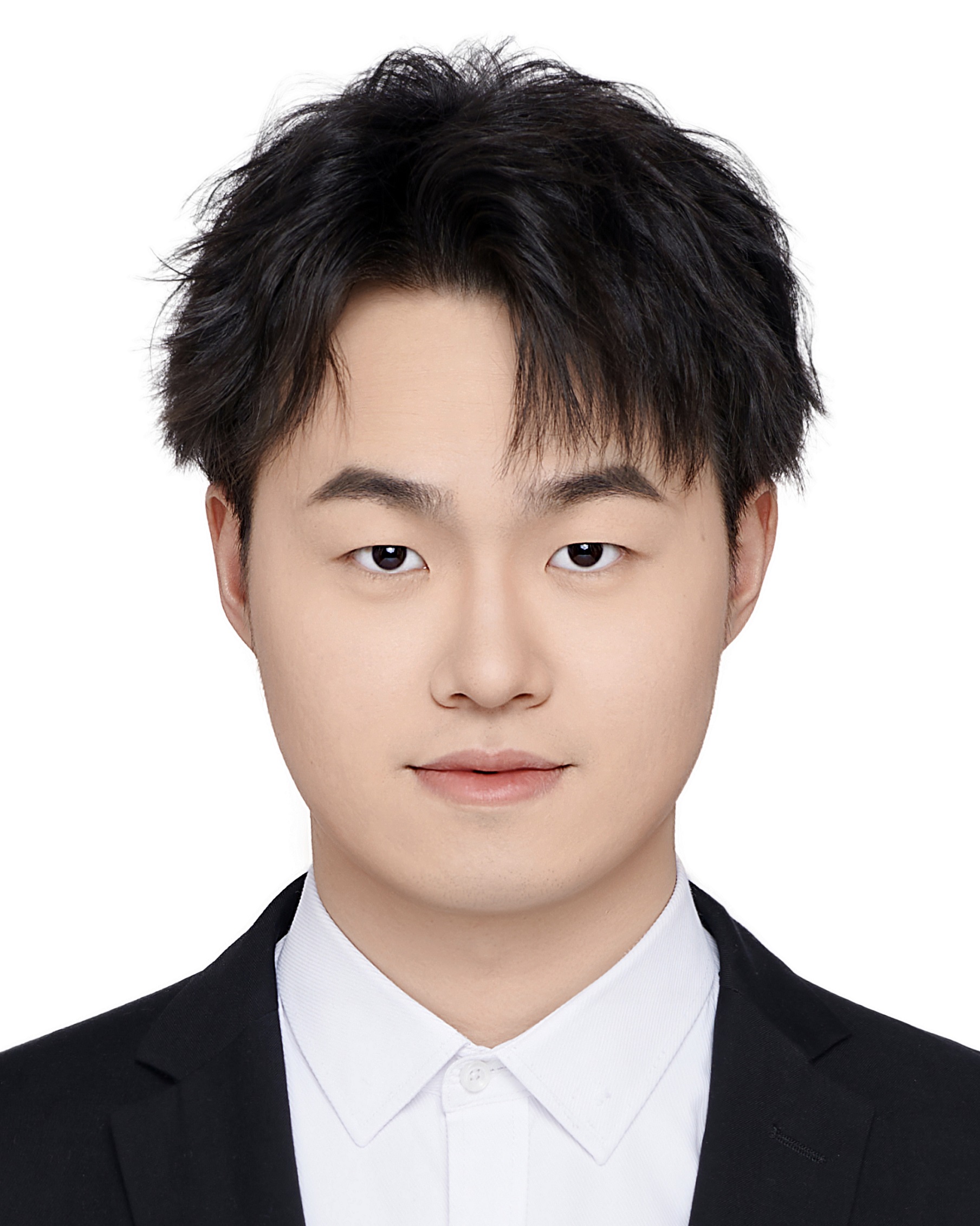}}]{Tong Nie (Student Member, IEEE)} received the B.S. degree from the college of civil engineering, Tongji University, Shanghai, China. He is currently pursuing dual Ph.D. degrees with Tongji University and The Hong Kong Polytechnic University. He has published several papers in top-tier avenues in the field of spatiotemporal data modeling, including KDD, AAAI, CIKM, IEEE TITS, and TR-Part C.
His research interests include spatiotemporal learning, time series analysis, and large language models.
\end{IEEEbiography}

\vspace{-20pt}
\begin{IEEEbiography}
[{\includegraphics[width=1in,height=1.25in,keepaspectratio]{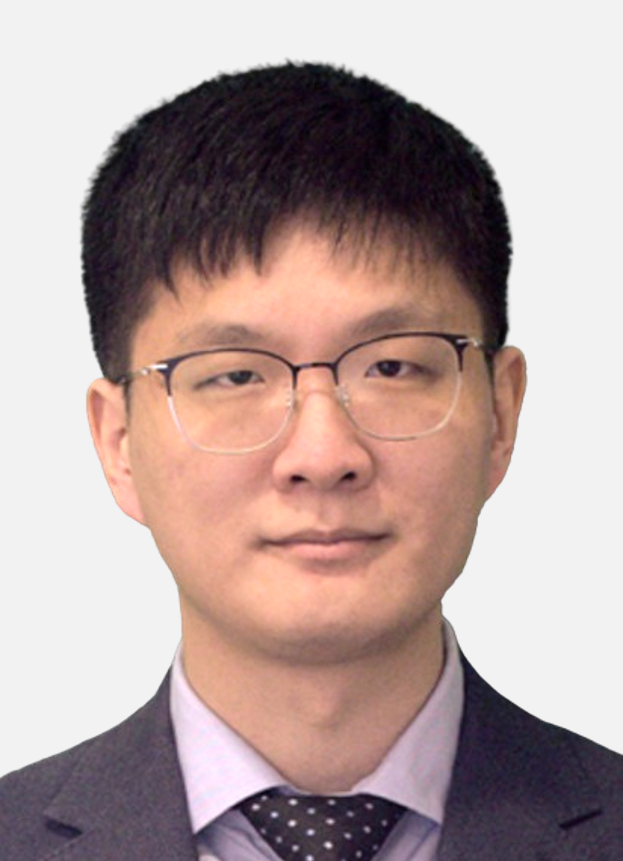}}]{Wei Ma (Member, IEEE)} received the bachelor’s
degree in civil engineering and mathematics from
Tsinghua University, China, and the master’s degree
in machine learning and civil and environmental engineering and the Ph.D. degree in civil and
environmental engineering from Carnegie Mellon
University, USA. He is currently an Assistant Professor with the Department of Civil and Environmental
Engineering, The Hong Kong Polytechnic University (PolyU). His current research interests include
machine learning, data mining, and transportation
network modeling, with applications for smart and sustainable mobility
systems.
\end{IEEEbiography}

\vspace{-20pt}
\begin{IEEEbiography}[{\includegraphics[width=1in,height=1.25in,keepaspectratio]{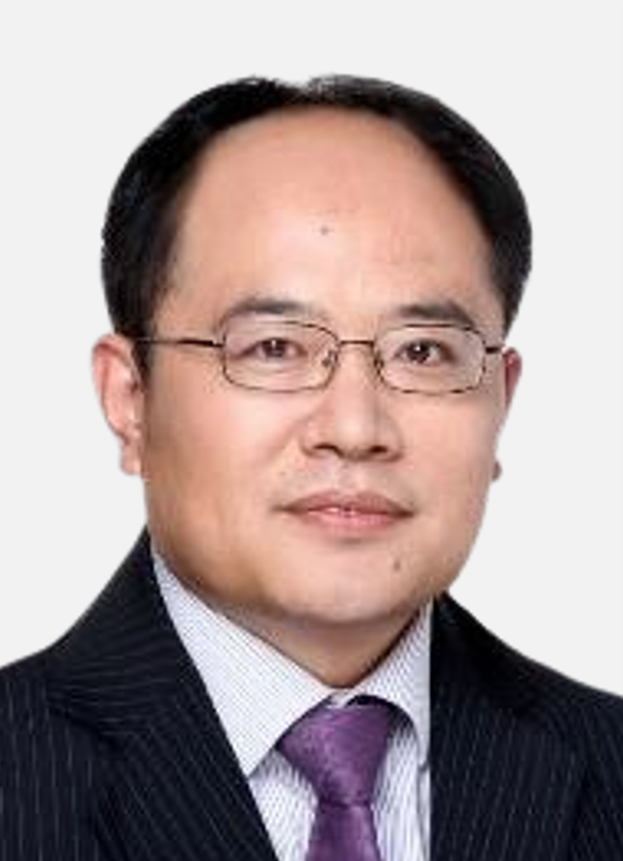}}]{Jian Sun} received the Ph.D. degree in transportation engineering from Tongji University, Shanghai, China. He is currently a Professor of transportation engineering with Tongji University. He has published more than 100 papers in SCI journals.
His research interests include intelligent transportation systems, traffic flow theory, AI in transportation, and traffic simulation. 
\end{IEEEbiography}

\vspace{-20pt}
\begin{IEEEbiography}[{\includegraphics[width=1in,height=1.25in,clip,keepaspectratio]{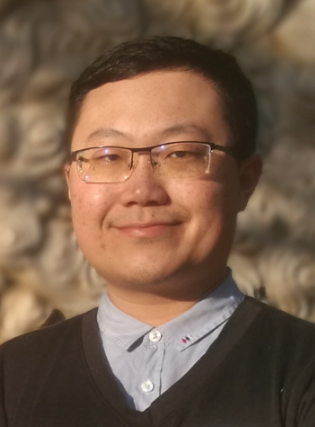}}]{Yu Yang} is currently an Assistant Professor with the Centre for Learning, Teaching, and Technology, The Education University of Hong Kong. He received the Ph.D. degree in Computer Science from The Hong Kong Polytechnic University in 2021. His research interests include spatiotemporal data analysis, representation learning, urban computing, and learning analytics.
\end{IEEEbiography}

\vspace{-20pt}
\begin{IEEEbiography}[{\includegraphics[width=1in,height=1.25in,clip,keepaspectratio]{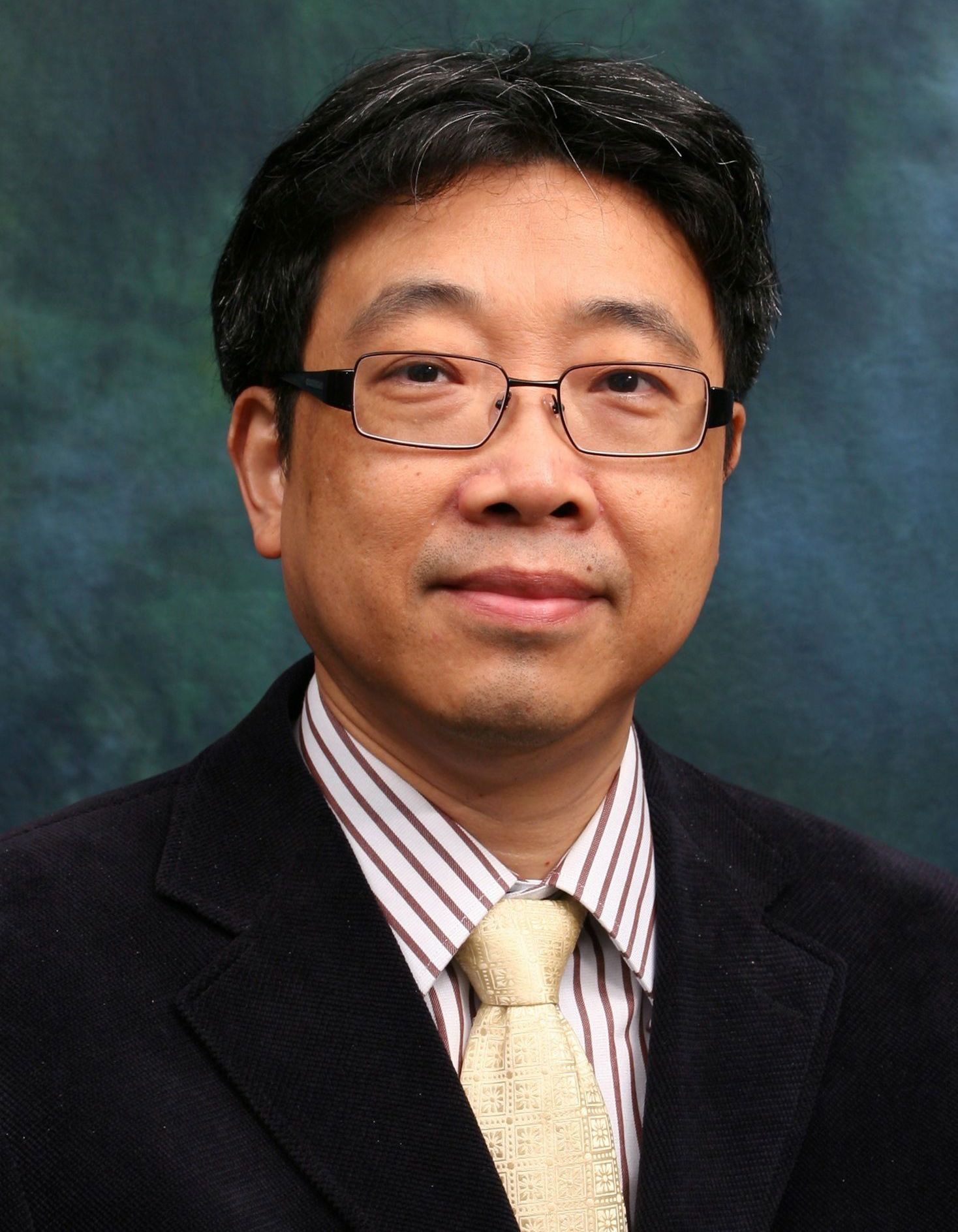}}]{Jiannong Cao (Fellow, IEEE)} received the M.Sc. and Ph.D. degrees in computer science
from Washington State University, Pullman, WA, USA, in 1986 and 1990, respectively. He is currently the Chair Professor with the Department of Computing, The Hong Kong Polytechnic University, Hong Kong. His current research interests include parallel and distributed computing, mobile computing, and big data analytics. Dr. Cao has served as a member of the Editorial Boards of several international journals, a Reviewer for international journals/conference proceedings, and also as an Organizing/ Program Committee member for many international conferences.
\end{IEEEbiography}

\end{document}